\documentclass[pdflatex,bst/sn-mathphys-num]{sn-jnl}

\usepackage{graphicx}%
\usepackage{multirow}%
\usepackage{amsmath,amssymb,amsfonts}%
\usepackage{amsthm}%
\usepackage{mathrsfs}%
\usepackage[title]{appendix}%
\usepackage{xcolor}%
\usepackage{textcomp}%
\usepackage{manyfoot}%
\usepackage{booktabs}%
\usepackage{algorithm}%
\usepackage{algorithmicx}%
\usepackage{algpseudocode}%
\usepackage{listings}%
\usepackage[acronym]{glossaries}
\usepackage{makecell}

\usepackage{hhline}
\usepackage{graphicx}
\usepackage[table]{xcolor}
\usepackage[dvipsnames]{xcolor}
\usepackage{pdfpages}
\usepackage{caption}
\usepackage{subcaption}

\theoremstyle{thmstyleone}%
\theoremstyle{thmstyletwo}%

\theoremstyle{thmstylethree}%

\makeglossaries

\begin{document}

\title[Article Title]{Beyond Cybathlon: On-demand Quadrupedal Assistance for People with Limited Mobility}


\author*[1]{\fnm{Carmen} \sur{Scheidemann}}\email{carmensc@ethz.ch}

\author[1]{\fnm{Andrei} \sur{Cramariuc}}\email{crandrei@ethz.ch}

\author[1]{\fnm{Changan} \sur{Chen}}\email{chencha@ethz.ch}

\author[1]{\fnm{Jia-Ruei} \sur{Chiu}}\email{jia-ruei.chiu@hexagon.com}

\author[1]{\fnm{Marco} \sur{Hutter}}\email{mahutter@ethz.ch}

\affil[1]{\orgdiv{ETH Zurich}, \orgname{Robotic Systems Lab}, \orgaddress{\city{Zurich}, \country{Switzerland}}}




\abstract{ 
\textbf{Background:} 
Assistance robots have the potential to increase the independence of people who need daily care due to limited mobility or being wheelchair-bound. Current solutions of attaching robotic arms to motorized wheelchairs offer limited additional mobility at the cost of increased size and reduced wheelchair maneuverability.

\textbf{Methods:} 
We present an on-demand quadrupedal assistance robot system controlled via a shared autonomy approach, which combines semi-autonomous task execution with human teleoperation. Due to the mobile nature of the system it can assist the operator whenever needed and perform autonomous tasks independently, without otherwise restricting their mobility. We automate pick-and-place tasks, as well as robot movement through the environment with semantic, collision-aware navigation. For teleoperation, we present a mouth-level joystick interface that enables an operator with reduced mobility to control the robot's end effector for precision manipulation.

\textbf{Results:} 
We showcase our system in the \textit{Cybathlon 2024 Assistance Robot Race}, and validate it in an at-home experimental setup, where we measure task completion times and user satisfaction. We find our system capable of assisting in a broad variety of tasks, including those that require dexterous manipulation. The user study confirms the intuition that increased robot autonomy alleviates the operator's mental load.

\textbf{Conclusions:}  
We present a flexible system that has the potential to help people in wheelchairs maintain independence in everyday life by enabling them to solve mobile manipulation problems without external support. We achieve results comparable to previous state-of-the-art on subjective metrics while allowing for more autonomy of the operator and greater agility for manipulation.

}




\maketitle

\section{Background}\label{sec:background}


\begin{figure*}[!t]
    \centering
    \begin{minipage}[b]{0.49\textwidth}
        \includegraphics[width=\textwidth]{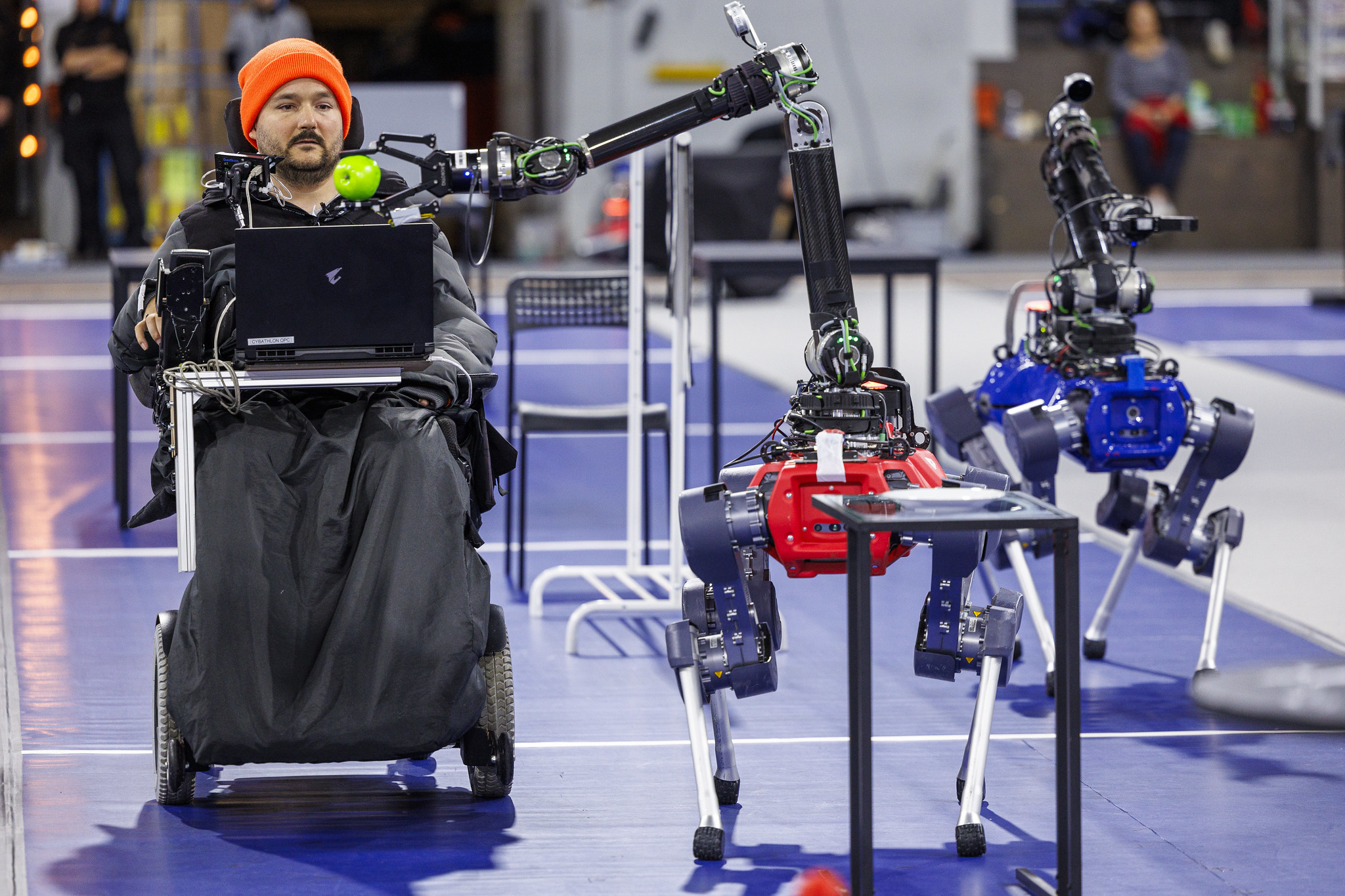}
      \end{minipage}
      \hfill
      \begin{minipage}[b]{0.49\textwidth}
        \includegraphics[width=\textwidth,trim={0 8.7cm 0 0},clip]{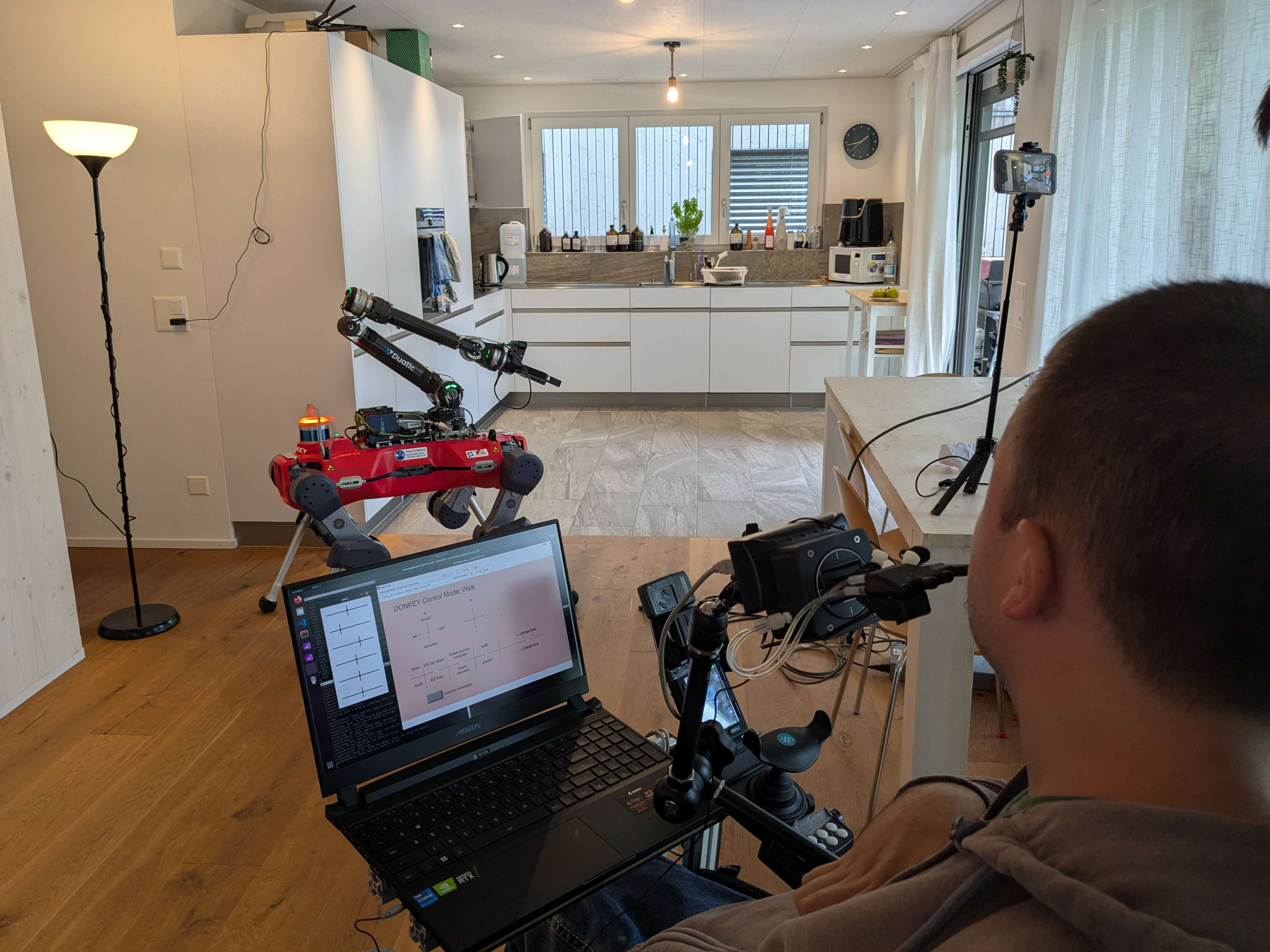}
      \end{minipage}
    \caption{\textbf{The proposed system in the two demonstrated environments:} (left) on the racetrack of the Cybathlon 2024 competition and (right) in the home of the operator.}
    \label{fig:header}
\end{figure*}

\subsection{Introduction}

People with severe motor impairments (from \textit{e.g.} muscular disease, spinal cord injury, cerebral palsy, or other neurological conditions) or missing limbs often struggle to perform everyday tasks independently. As a result, they typically rely on dedicated assistance from nurses or other caretakers. This limits the independence of the affected people, which is an essential aspect of health and happiness~\cite{doi:10.1080/09638280110072922}. Assistive robotic technologies offer a promising avenue for restoring independence~\cite{doi.org/10.1155/2012/538169} and are increasingly investigated for applications in rehabilitation~\cite{7967538}, personal assistance~\cite{9223470}, and social support~\cite{Chapter21Companionrobotsforwellbeingareviewandrelationalframework}.
More recently, the potential of generalist humanoid assistance robots has received broader attention~\cite{carnevale2025exploring, laban2024social}; however, technical limitations and restricted capabilities continue to hinder their practical deployment~\cite{maure2025autonomy}.
As such, most existing solutions for electric wheelchair users focus on robotic arms mounted directly onto the wheelchair~\cite{doi:10.1080/17483107.2021.2017030, 7814381, AssistiveInnovation}. These mounted systems face inherent trade-offs: the added size and weight reduce maneuverability, while sharing the wheelchair’s battery constrains power usage and operating time. As a result, such arms are typically limited in payload and reach: for instance, the commercial iARM can lift only 1.5kg~\cite{10.1097, AssistiveInnovation}. Additionally, scaling up the hardware either restricts mobility or reduces the usable workspace. This gap underscores the need for alternative assistive platforms that maintain mobility while providing robust manipulation capabilities.

Our proposed solution is the use of an independent quadrupedal robotic platform, ANYmal~\cite{7758092}, with a robotic arm, DynaArm~\cite{Dynaarm}, mounted on top. This setup offers more versatility than mounting an arm directly on the wheelchair, as the robot can reach places the wheelchair cannot, has a separate power supply, and does not limit the wheelchair's maneuverability or bloat its size. The independent robot can perform tasks autonomously without the operator's direct supervision, \textit{e.g.} fetch things from another room or empty a dishwasher while the operator does something else. It also poses less of a safety risk to the operator, as system failures are less likely to affect the wheelchair's basic mobility, and it does not require the operator to carry a larger battery or remain in constant proximity to a moving robotic arm. Despite these advantages, research on utilizing mobile manipulators as assistive devices remains scarce~\cite{mivseikis2020lio, bilyea2017robotic, cabrera2021exploration, nanavati2023physically}. Mišeikis \textit{et al.} and Bilyea \textit{et al.} present autonomous systems designed for a restricted subset of care-related tasks~\cite{mivseikis2020lio, bilyea2017robotic}, while Cabrera \textit{et al.} introduce a mobile manipulator teleoperated by an external caregiver~\cite{cabrera2021exploration}. Although these works demonstrate the feasibility of mobile robotic assistance, none of them investigate a system that can be independently operated by the affected person to provide general support in activities of daily living.

One notable exception is the work of Padmanabha \textit{et al.}~\cite{padmanabha2024independence}, who propose a teleoperated mobile manipulator to assist a person with quadriplegia in everyday tasks. Their system is controlled via a wearable head-mounted device (HAT) that maps head movements to robot commands, and is evaluated in an at-home environment across a wide range of daily activities. The authors report successful task execution using both pure teleoperation and a shared-autonomy mode termed “Driver Assist,” demonstrating the practical potential of direct user control.
While inspiring, the proposed approach exhibits two key limitations that our work addresses: Firstly, the operator must manually navigate the robot to any item they wish to interact with, while avoiding collisions with the environment. This can be a challenging task from a single, fixed viewpoint and often requires the operator's full attention. Secondly, the selected robot embodiment, the \textit{Hello Robot Stretch}, imposes inherent physical constraints. The limited payload and manipulation strength restrict the system to lightweight objects, such as tissues, playing cards, or beverage cans, and preclude essential tasks such as opening doors or transporting heavier items. Furthermore, the wheeled base cannot traverse complex terrain, keeping the assistant inherently limited to a home application. In contrast, a legged robotic base is almost limitless in the terrain it can overcome, effectively extending the range of space the operator can access. For example, it would enable them to have the robot fetch something from atop a flight of stairs.

Bringing more attention to the topic of assistive technology is one of the main goals of Cybathlon~\cite{jaeger2023cybathlon, Riener2016}, which is a competition dedicated to showcasing and advancing technologies that assist people with disabilities, \textit{e.g.}, individuals who are blind, those in wheelchairs, or those with prosthetic arms. The \textit{Robotic Assistance Race} (ROB) of Cybathlon 2024 focused on robotic assistance platforms for individuals with severe motor impairments such as quadriplegia. The competing platforms took the form of either an arm attached to the wheelchair or separate moving robots that the operator can remotely control. The race consisted of a set of everyday tasks that the operator had to complete on their own, utilizing only the robotic assistant. Examples of the tasks completed in the course of the race include: collecting a package from a mailbox, brushing the operator’s teeth, hanging a scarf on a clothesline, and emptying a dishwasher.

Our evaluation of the system, visualized in Figure~\ref{fig:header}, is two-fold: Firstly, we demonstrate its capabilities at the \textit{Cybathlon} 2024 competition, achieving a competitive ranking and completing nine out of the ten tasks successfully within a strict time limit. Second, we validate the real-world applicability of our system in the operator's home, a much more realistic and unstructured environment than the Cybathlon racetrack. Nonetheless, our system successfully solves various tasks even on the first attempt. Examples of the completed tasks include warming up a meal in a microwave and pouring a glass of water before inserting a straw. We demonstrate both the system's dexterity in manipulation-only tasks and its versatility in semi-autonomous task execution. We evaluate operator experience and mental load using a subjective questionnaire and compare the completion times of specific tasks with those of similar works that solve the same objective.

\section{Methods}\label{sec:method}






\vspace{-5pt}
\begin{figure*}[h]
    \centering
    \includegraphics[width=0.98\textwidth]{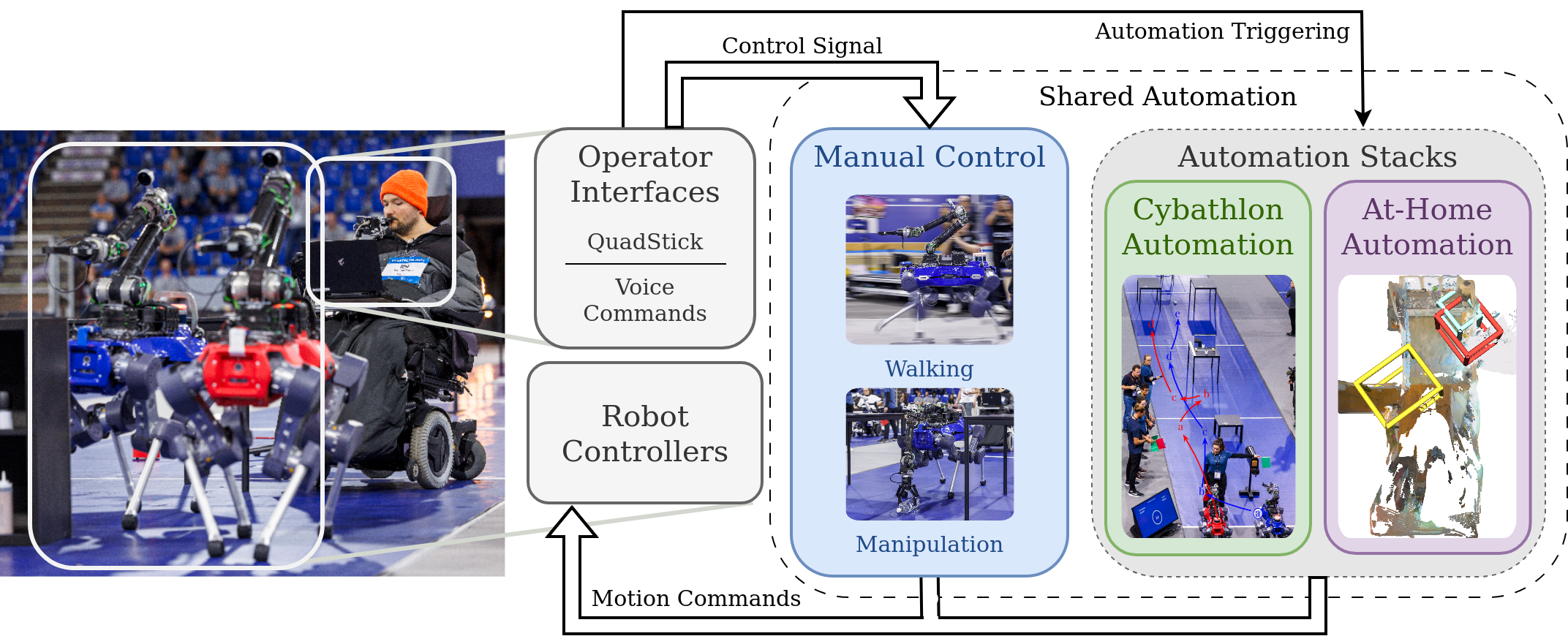}
    \vspace{2pt}
    \caption{\textbf{An overview of the proposed system}:
    The system, as presented both at the Cybathlon 2024 competition and in an at-home environment, has two central pillars, which together form the shared autonomy stack. The first pillar, the Manual Control stack (in blue), describes our teleoperation interface. We take the control signal from the operator joystick, the Quadstick, and translate it into control commands for the robot's walking or manipulation controllers. The second pillar, the Automation stack (in green and purple), is application-specific. It handles the autonomous execution of pre-defined tasks after receiving an activation signal from the operator interfaces. The robot can only be in one state at a time, i.e., it is being fully teleoperated or moving fully autonomously. Pictured on the left are two robots and the expert operator, getting into position to solve a task during Cybathlon 2024.
    }
    \label{fig:system}
\end{figure*}

\subsection{System}
The goal of this work is to create a system that can support an operator with limited limb dexterity in their daily activities at home, as well as excel in the Cybathlon competition. 
The nature of Cybathlon as a time-limited race necessitates a robust and reliable method for solving complex manipulation tasks, such as unloading a dishwasher. Automating such interactions robustly is the subject of much current cutting-edge robotics research~\cite{zhang2021safe, yin2021modeling, kroemer2021review, torne2024reconciling}.
Due to the nature of machine learning these methods struggle in out-of-distribution cases. Additionally, the current limitations of vision sensors make certain object types, such as those that are very thin or transparent, almost impossible to autonomously grasp.
To address this, we additionally propose an intuitive, easy-to-use teleoperation system designed for people with limited mobility. However, long-horizon teleoperation of a complex robotic system places a significant mental burden on the operator~\cite{doi:10.1177/1729881419888042}. To mitigate this, we integrate various autonomy modules that take over when fine-grained manual control is not required. The entire pipeline is illustrated in Figure~\ref{fig:system}. We denote this combination of manual teleoperation and autonomous task execution as shared autonomy. Both the Cybathlon and the At-Home Automation stack present independent solutions to the same problem: pre-positioning the robot in front of objects the operator wants to interact with. As such, they are interchangeable within the system. This split of labor between teleoperation and automation not only alleviates the operator's mental load, but also opens the door to more complex competition strategies, such as the \textit{Two-Robot Solution} described in Section \ref{sec:cybathlon_auto}.

Both system stacks are operated through interfaces that require no limb dexterity from the user. The primary control device is a QuadStick, a gaming-controller-style interface positioned at the operator’s mouth. It is complemented by a voice interface that allows the operator to issue higher-level verbal commands. Both interfaces can trigger elements of the automation stack, with the choice of input modality tailored to the application context. During \emph{Cybathlon}, where speed and reliability are critical, the voice interface is omitted as it would mean relying on a stable internet connection on the racetrack, as well as introducing wait times due to the online ChatGPT \cite{chatgpt} API queries.
In the \emph{At-Home} setting, however, natural and flexible interaction is prioritized, making voice input valuable for specifying targets in an open-vocabulary manner. Across both settings, fine-grained manual control is performed exclusively via the QuadStick to ensure maximal precision.
To avoid conflicting commands, we enforce that only one stack is active at any given time, during which it sends pose commands directly to the robot, controlling either its base or the arm mounted on top.

The proposed system thus leverages the advantages of human teleoperation and cutting-edge automation to efficiently complete a wide variety of common activities of daily living. In the following sections, we describe the system in full, providing details on all aspects of its implementation, from hardware choices to control scheme mappings.

\subsubsection{Hardware}

\begin{figure*}[h]
    \centering
    \includegraphics[width=0.9\textwidth]{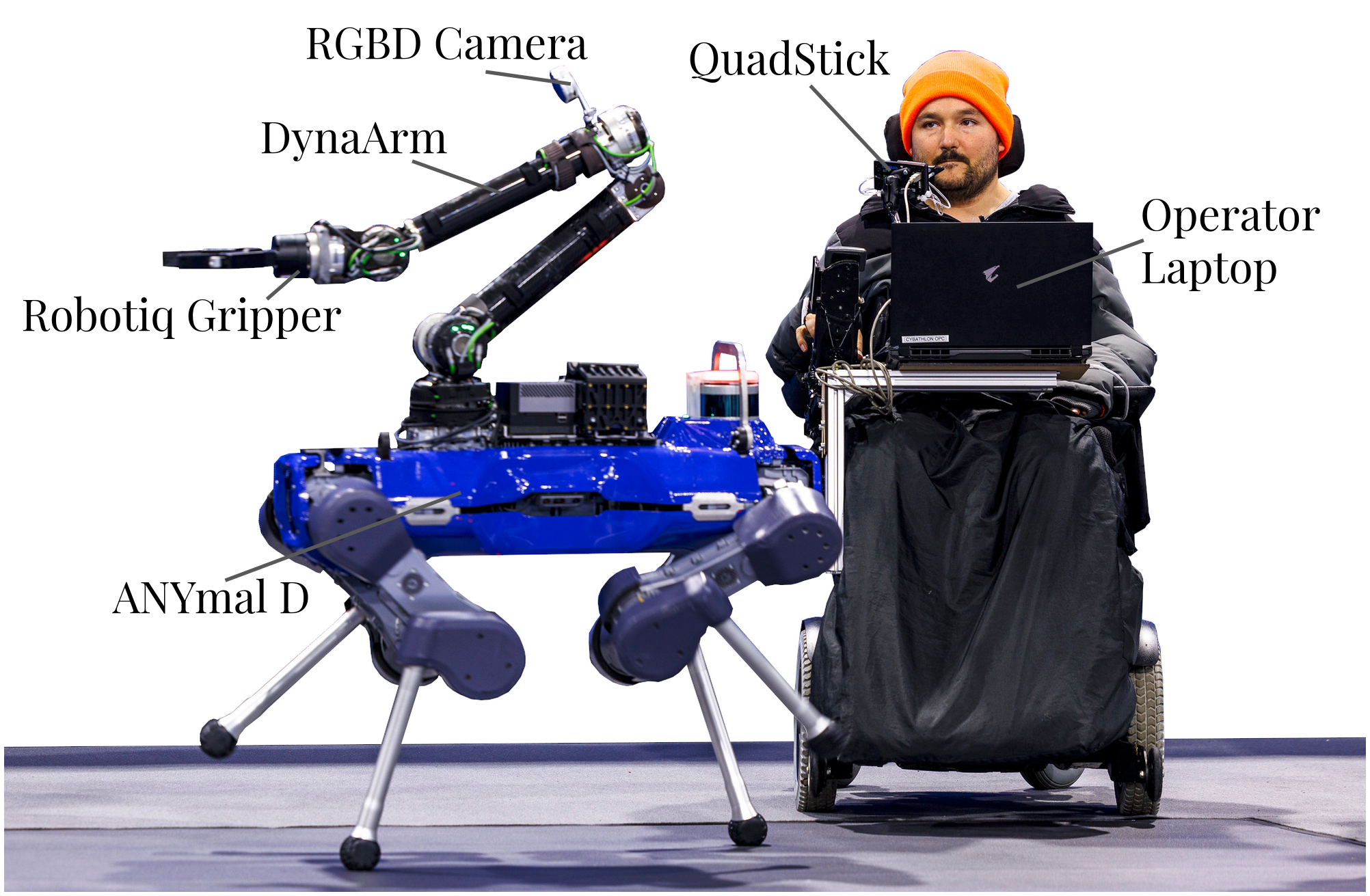}
    \caption{\textbf{An overview of the hardware involved in the proposed system}, in operation during \textit{Cybathlon 2024}. The main components are an ANYmal D quadrupedal base and a Duatic DynaArm robotic arm, to which we attach a Robotiq 140F Gripper and an Intel RealSense L515 camera. Attached to the wheelchair of the operator are the QuadStick, which enables them to steer the robot, and the Operator Laptop, which relays critical information between the robot and operator.
    }
    \label{fig:embodiment}
\end{figure*}

An overview of the hardware components of the system is shown in Figure~\ref{fig:embodiment}.
The robot itself is a combination of an ANYbotics ANYmal D body, which serves as the quadrupedal base, and a Duatic DynaArm~\cite{Dynaarm}, for manipulation. This platform allows both rapid traversal of the racetrack, at a maximum walking speed of $1.3$ m/s, and manipulation of heavy objects, such as doors, with sufficient motor torque to achieve a continuous carrying capacity of 6 kg.
The ANYmal body is largely unmodified from its original configuration, with the main differences being the custom payload, which consists of an additional GPU-equipped computer for high-bandwidth GPU processing (Jetson Orin) and the robotic arm. 
The arm is additionally equipped with a parallel gripper end-effector (Robotiq 2F-140), which allows it to grasp large items, as well as an on-board RGB-D camera, situated at the elbow similar to prior works~\cite{tulbure2024fast, tagmap}, providing the vision for automation tasks. The operator laptop is an Aorus 15P, to which we outsource some of the low-bandwidth GPU compute of the stack.
Together, these components form a versatile system that can move quickly, take on a wide range of manipulation tasks, and provide enough onboard computational power to enable key automation.

\subsubsection{Low-Level Robot Controllers}

A low-level robot controller takes higher-level commands (e.g., moving the arm or foot to a target position) and transforms them into motor-level commands to achieve the task. To enable both precise manipulation and efficient locomotion, the system relies on two complementary controller approaches that operate under distinct paradigms. They were both developed prior to this project and applied as is, with minimal parameter retuning. The first is OCS2~\cite{9387121}, a \acrfull{mpc}-based controller used for precise, 6 \acrfull{dof} stable full-body control of the end-effector. Although the controller is not used for locomotion, it has access to the full system, allowing the base to tilt and translate while keeping all feet in contact with the ground. This significantly extends the robot's workspace. For example, the base can pitch down when retrieving a fallen bottle from the floor or lean forward to grab an item placed far away, extending its reachable workspace. Additionally, full-body control of the system enables the robot to autonomously avoid self-collision between the arm and the base, easing the control for the operator. Compared to learning-based full-body controllers~\cite{fu2023deep}, its model-based nature enables it to maintain stable, high-precision end-effector pose control. However, the \acrshort{mpc} controllers' locomotion capabilities are limited. Therefore, movement from one point to the next is achieved through our second control method: a learned policy trained in simulation using reinforcement learning \cite{doi:10.1126/scirobotics.abk2822}. The policy operates in 3 \acrshort{dof}, accepting $x$, $y$, and $yaw$ targets for the base. These two controllers are separate entities that the operator, using the QuadStick, can seamlessly switch between to either move the base or end-effector of the robot. As such, these control modes provide a flexible way to command the robot’s motion, allowing smooth transitions between fine end-effector control and agile base movement.

\subsubsection{Physical Operator Interfaces}
The physical interfaces allow the operator to send commands to the robot as shown in Figure~\ref{fig:system}, which ultimately result in motor commands produced by the aforementioned low-level robot controllers. To ensure accessibility, we present a specialized interface that is non-reliant on operator limb dexterity.
The basis of our main operator interface is a QuadStick gaming controller, which the operator can interact with using their mouth. This controller sits in front of the operator's face and was specifically designed to enable people with limited mobility to play video games on conventional gaming consoles. The QuadStick controls will be discussed in depth in this and the following section. The operator maneuvers themselves using their personal motorized wheelchair, to which we attach both a display laptop and the QuadStick. The display laptop relays important system information, such as camera images from the robot, to the operator, and receives high-level input from the operator via the QuadStick's point-and-click capabilities or a voice command interface. Together, the QuadStick and display laptop allow the operator to receive system feedback, issue high-level commands, or send a continuous control signal to the robot. The wheelchair control itself remains unmodified from the everyday of the operator: control is facilitated through a joystick which they control with their wrist. It was an active design choice within our interfaces to not interfere with this mechanism so that we don't limit their mobility.

\subsubsection{Manual Control Mappings}

To reliably complete dexterous manipulation tasks, an intuitive teleoperation interface is crucial. Thus, one of our core contributions is the operator's manual control interface, previously visualized as the \textit{Manual Control} stack in Figure~\ref{fig:system}. The following describes the exact control mapping in detail. It was developed via multiple feedback cycles in collaboration with our competition operator in the months leading up to Cybathlon 2024. 

The physical interface of our primary input device, the QuadStick, is fairly limited and consists of a mouthpiece and a push button mounted on a two-axis joystick, which the operator can control by moving the mouthpiece along either axis with their lips. The mouthpiece itself contains four breath-based input channels, each of which can register no input, blowing, or sucking, in a binary manner. Of the four channels, we may only use three for robotic control, as the fourth instead controls the internal mode switching of the device. The difficulty here lies in reducing control complexity without compromising performance. The robots each possess nine \acrlong{dof}, three in the base and six in the arm. These nine \acrshort{dof}, as well as the switching between base and end-effector controllers, activating the gripper, and switching between robots, need to be mapped down to the seven available input channels and the primary and secondary axis mappings of the QuadStick. 

\begin{figure}[h]
    \centering
    \includegraphics[width=13cm]{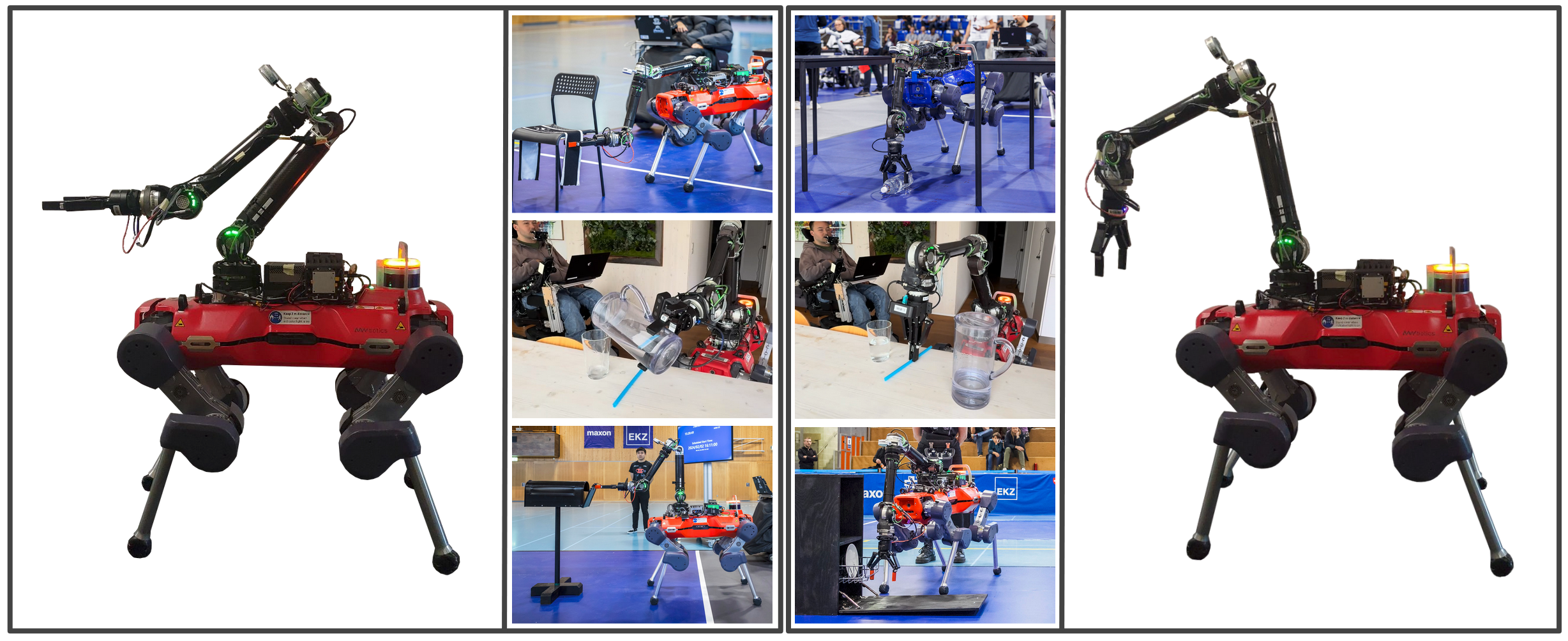}
    \caption{\textbf{The initial configurations of the two \acrlong{ee} Control Modes.} The left of the Figure shows the initial configuration of \textit{\acrshort{ee} Control Mode Front}, the right that of \textit{\acrshort{ee} Control Mode Top}. Examples of how the control modes can be used are visible in the two central columns. For \textit{\acrshort{ee} Control Mode Front}, these include picking up a scarf from a chair (at the \textit{Cybathlon 2024 February Challenge}), pouring water from a carafe into a glass (at home), and opening a mailbox (at the \textit{Cybathlon 2024 February Challenge}). For \textit{\acrshort{ee} Control Mode Top}, they include picking up a large bottle of water from the floor (at \textit{Cybathlon 2024}), picking up a straw from a table (at home), and pulling the rack out of a dishwasher (at the \textit{Cybathlon 2024 February Challenge}).}
    \label{fig:control_modes}
\end{figure}

To make the end-effector control more intuitive, we implemented two different control modes within our \acrshort{mpc}-based controller~\cite{9387121}, which the operator can toggle between. Their core difference lies in their initial arm configuration, as shown in Figure~\ref{fig:control_modes}, along with examples of their application. The first control mode is referred to in Table~\ref{table:axis_mapping} as \textit{\acrshort{ee} Control Mode Front} and is appropriate for most tabletop pick-and-place tasks, including bulky objects, such as a carafe, or those demanding high dexterity, such as a scarf.
The second is referred to as \textit{\acrshort{ee} Control Mode Top} and is best suited for manipulation tasks near the floor, such as unloading a dishwasher. 
The third control mode implemented is referred to as \textit{Base Control Mode}, in which the operator can walk the robot from one location to the next, using the robust locomotion policy. The following sections detail the input mappings for each mode, beginning with those shared across modes and subsequently describing the mode-specific mappings.

\begin{figure}[h]
    \centering
    \includegraphics[width=12cm]{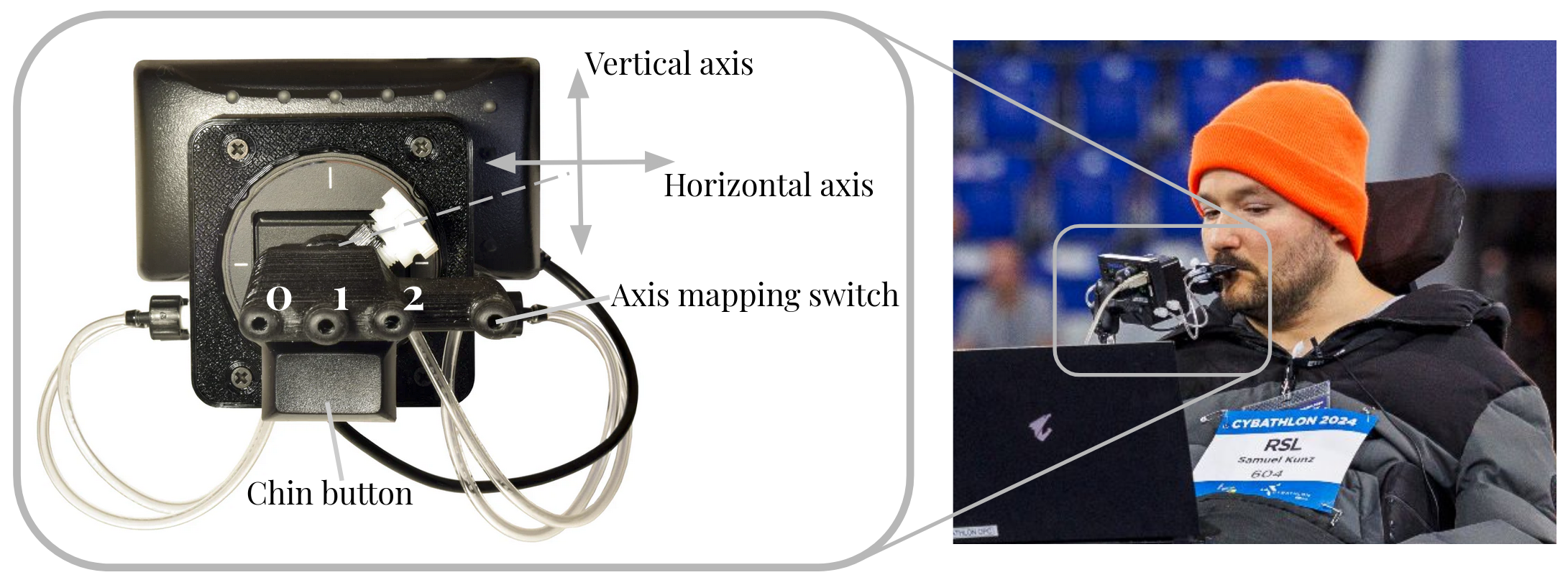}
    \caption{\textbf{The layout of the QuadStick input options} is pictured on the left: the central two-axis joystick encoding the vertical and horizontal axes, the three central control channels (marked as 0, 1, and 2), the axis mapping switch control channel to the right, and below them the chin button. On the right, the operator, Samuel Kunz, is pictured using the QuadStick at \textit{Cybathlon 2024}.}
    \label{fig:quad_vis}
\end{figure}

\subsubsection*{Mode-Independent Inputs}

The left-most breath-based input channel (labeled as Channel 0 in Figure~\ref{fig:quad_vis}) is used to switch between the two \acrlong{ee} control modes. The middle-right input channel (Channel 2 in Figure~\ref{fig:quad_vis}) controls the opening and closing of the gripper on suck and switches to the base control mode on blow. The right-most input channel (labeled \textit{Axis mapping switch}) cannot be used directly for control, as it is used internally in the QuadStick firmware to switch between the primary and secondary joystick axis mappings, or to switch into the tertiary one, which accesses \acrshort{gui} control of the laptop. The \textit{chin button} was used to switch the active control between the two robots during the Cybathlon 2024 competition. As the at-home experiments involved only a single robot, we instead map this to a safety override: pressing it cancels all active execution on the robot and disconnects the operator control signal. The operator can press it again later to regain control of the robot, if desired.

\subsubsection*{Mode-Specific Inputs}
\label{sec:axis_mapping}

Although the QuadStick has only one physical joystick, it is possible to switch between two axis assignments, denoted here as primary and secondary joystick axis mapping modes, using the rightmost breath input channel. Repeated mode switching would substantially slow down control, as this switch takes a few seconds to complete. In the at-home application, this would only be bothersome; however, during a competition such as Cybathlon, time efficiency is crucial. Therefore, the mappings are designed so that the majority of movement can be completed using the primary joystick axis mapping, and the secondary axis is only needed for final adjustments to the manipulation angle.

\begin{figure}
\centering
\begin{subfigure}{.6\textwidth}
  \centering
  \begin{tabular}{|l|l|l|ll|}
        \hline
        \begin{tabular}[c]{@{}l@{}}Mapping\\ Mode\end{tabular} &
        \multicolumn{1}{l|}{Axis} &
        \multicolumn{1}{l|}{\begin{tabular}[c]{@{}l@{}}Base\\Control\\Mode\end{tabular}} &
        \multicolumn{1}{l|}{\begin{tabular}[c]{@{}l@{}}EE \\ Front\\Mode\end{tabular}} &
        \begin{tabular}[c]{@{}l@{}}EE \\ Top\\Mode\end{tabular} \\ \hline
        \multirow{2}{*}{Primary} & horizontal & \multicolumn{3}{c|}{x} \\ 
        \hhline{|~|----|}
         & vertical & \multicolumn{3}{c|}{y} \\ \hline
        \multirow{2}{*}{Secondary} & horizontal & \multicolumn{1}{c|}{-} & \multicolumn{2}{c|}{roll} \\ 
        \hhline{|~|----|}
         & vertical & \multicolumn{1}{c|}{-} & \multicolumn{1}{c|}{yaw} & \multicolumn{1}{c|}{pitch} \\ \hline
        \end{tabular}
  \caption{The axis mappings of the QuadStick joystick.}
  \label{table:axis_mapping}
\end{subfigure}%
\begin{subfigure}{.4\textwidth}
  \centering
  \includegraphics[width=0.6\linewidth]{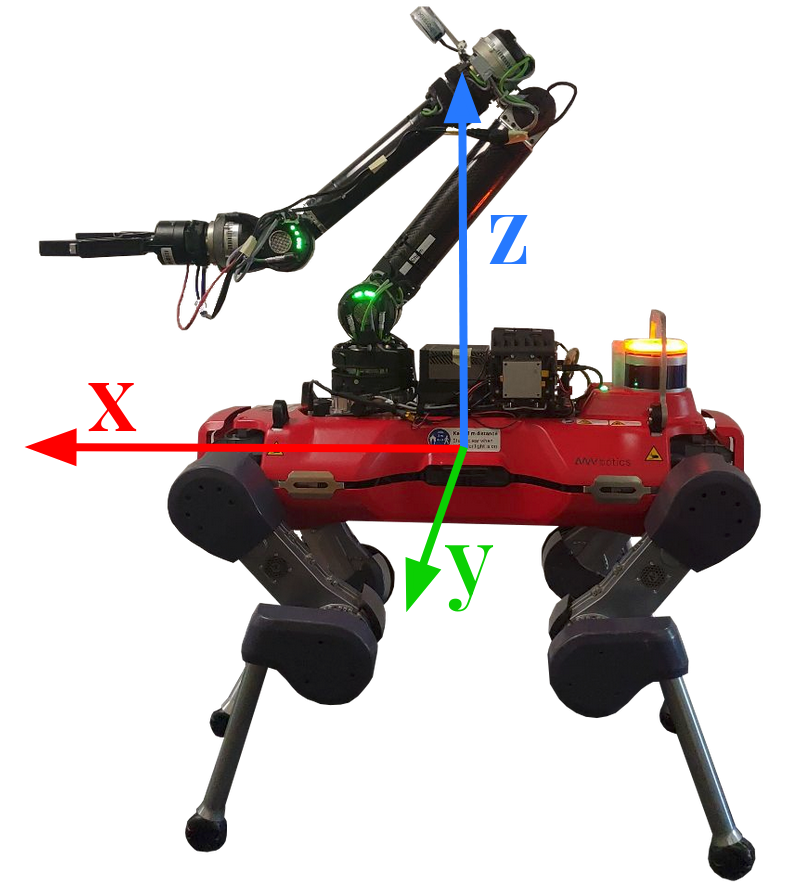}
  \caption{The robot frame axes.}
  \label{fig:axes}
\end{subfigure}
\caption{\textbf{The primary and secondary mappings of the horizontal and vertical axes of the joystick of the QuadStick.} The axes are defined within the fixed frame of the robot, with the origin in the center of its body; the x-axis corresponds to the axis pointing in the nominal walking direction of the robot (the forward direction), the y-axis corresponds to the perpendicular sideways direction, and the z-axis points upwards. Not pictured is the tertiary axis mapping, which provides a point-and-click trackpad interface to the operator laptop.}
\label{fig:test}
\end{figure}

To keep control as intuitive for the operator as possible, the primary mapping is shared over all modes. It maps the horizontal and vertical axes to $x$ and $y$ movement of the base or end-effector, respectively. The axis definitions are static within the robot base and visualized in Figure~\ref{fig:axes}. The middle-left channel (referred to as channel 1 in Figure~\ref{fig:quad_vis}) is mapped in a controller-specific manner. In all modes, channel 1 maps the third control axis of the currently active controller, as the first two are being mapped via the joystick. For the base control mode, this corresponds to the yaw axis. For end-effector control, we map this channel to the z-axis; thus, all translational axes are exposed when in primary mapping mode.

The secondary axis mappings are chosen to expose the two most useful rotational axes for each specific control mode. These were identified through an iterative testing process with the Cybathlon operator and are shown in Table \ref{table:axis_mapping}. Both the \textit{front} and \textit{top} configurations share a mapping of roll onto the horizontal axis. However, the vertical axis maps to yaw in \textit{\acrshort{ee} Control Mode Front}. This is useful, for example, to reach around an obstacle and grasp an item from an angle. In \textit{\acrshort{ee} Control Mode Top}, the vertical axis instead maps to pitch. This axis is necessary, for example, to place a plate, which was vertically lifted from a rack, down horizontally. The secondary mode remains unmapped for the base controller, as this controller has three degrees of freedom rather than six.

Although this control scheme was initially developed for Cybathlon 2024, it encodes control motions universal to many activities of daily life, such as lifting objects from the ground or from elevated surfaces, or turning knobs. The generality of the scheme is underscored by its direct applicability from Cybathlon to the at-home scenario, without any modifications. 

\subsection{The Cybathlon 2024 Competition}
\label{sec:comp_desc}
The Cybathlon is an international competition that brings together teams from around the world to demonstrate cutting-edge assistive technologies, which help people with physical disabilities in everyday life. Unlike traditional sports, the focus is not solely on athletic performance, but also on how well these advanced devices (such as prosthetic limbs, exoskeletons, and assistive robots) support the athletes in simulated real-life tasks. To be precise, the competition consists of a series of races, each focused on assisting with a specific disability. 

\begin{figure*}[h]
    \centering
    \includegraphics[width=\textwidth]{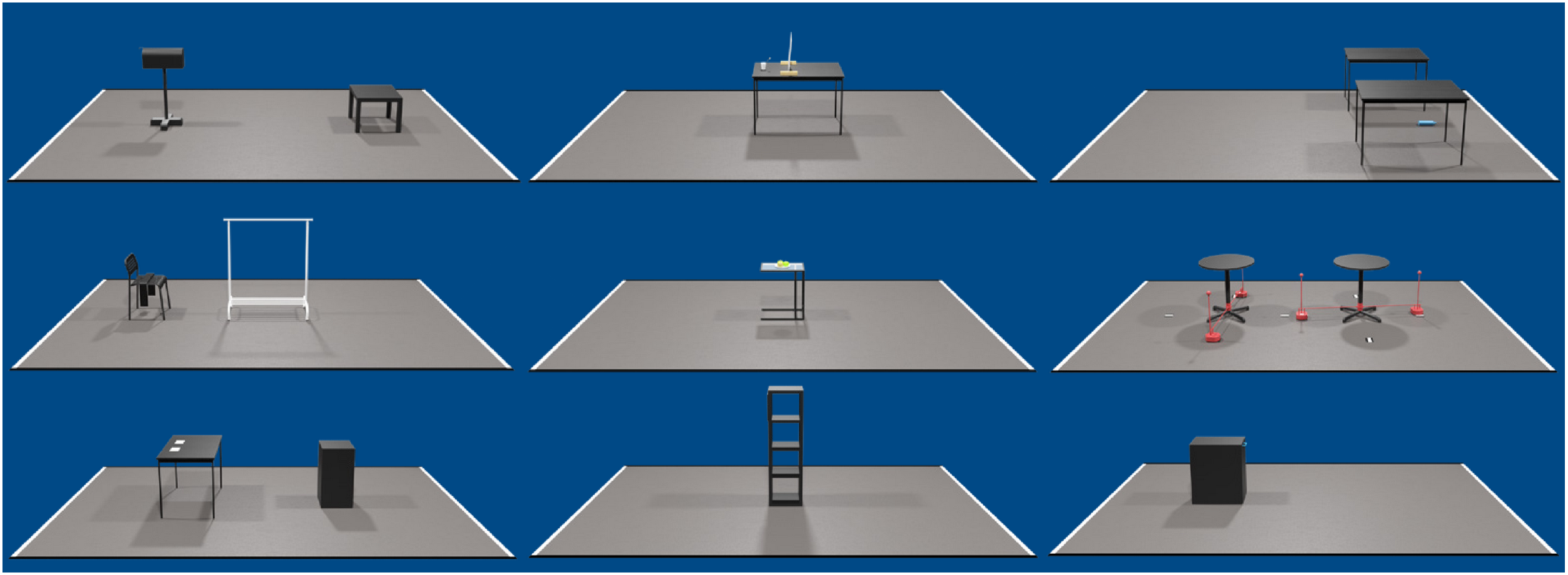}
    \caption{\textbf{The tasks forming the \textit{Assistance Robot Race}}: From left to right, top to bottom: \textit{Mailbox, Toothbrush, Pick Up, Scarf, Eating, Crowd, Spice Up, Touchscreen, Dishwasher}. Not pictured is the \textit{Door} task, the only task which was not attempted in any of the authors' team's runs.}
    \label{fig:track}
\end{figure*}

We showcased the presented system in the \textit{Assistance Robot Race}, a race designed around helping people with severe motor impairments (wheelchair users with a severe impairment of both upper limbs) overcome everyday obstacles. The race consists of ten tasks that must be solved successively in under ten minutes. Specifically, Figure~\ref{fig:track} presents the tasks in this category, which are described as follows: (I) \textbf{Mailbox:} Open a mailbox and retrieve a package, (II) \textbf{Toothbrush:} Pick up a toothbrush from a cup, touch the operator's mouth with it, and return it, (III) \textbf{Pick Up:} Pick up a large bottle of water from the floor and place it on a table, (IV) \textbf{Scarf:} Pick up and hang a scarf on a clothesline, (V) \textbf{Eating:} Grab an apple from a plate, bring it to the operator's mouth, and return it, (VI) \textbf{Crowd:} Pass between a series of moving obstacles without touching them, (VII) \textbf{Spice Up:} Grab two different specific spices, specified at runtime, from a selection on a shelf and place them at a set location, (VIII) \textbf{Door:} Open a door for the operator and pass through after him, (IX) \textbf{Touchscreen:} navigate a touchscreen to select a predefined item, (X) \textbf{Dishwasher:} open a dishwasher and remove a plate from within. The Door task was the only task we did not attempt in any of the competition runs; the reasoning for this will be described in the following chapter.
The team additionally participated in the \textit{February 2024 Cybathlon Challenge}, a pre-competition consisting of a subset of the final tasks, namely \textit{Mailbox, Toothbrush, Scarf,} and \textit{Dishwasher}.

\subsubsection{Cybathlon Automation}
\label{sec:cybathlon_auto}

\begin{figure*}[h]
    \centering
    \includegraphics[width=0.95\textwidth]{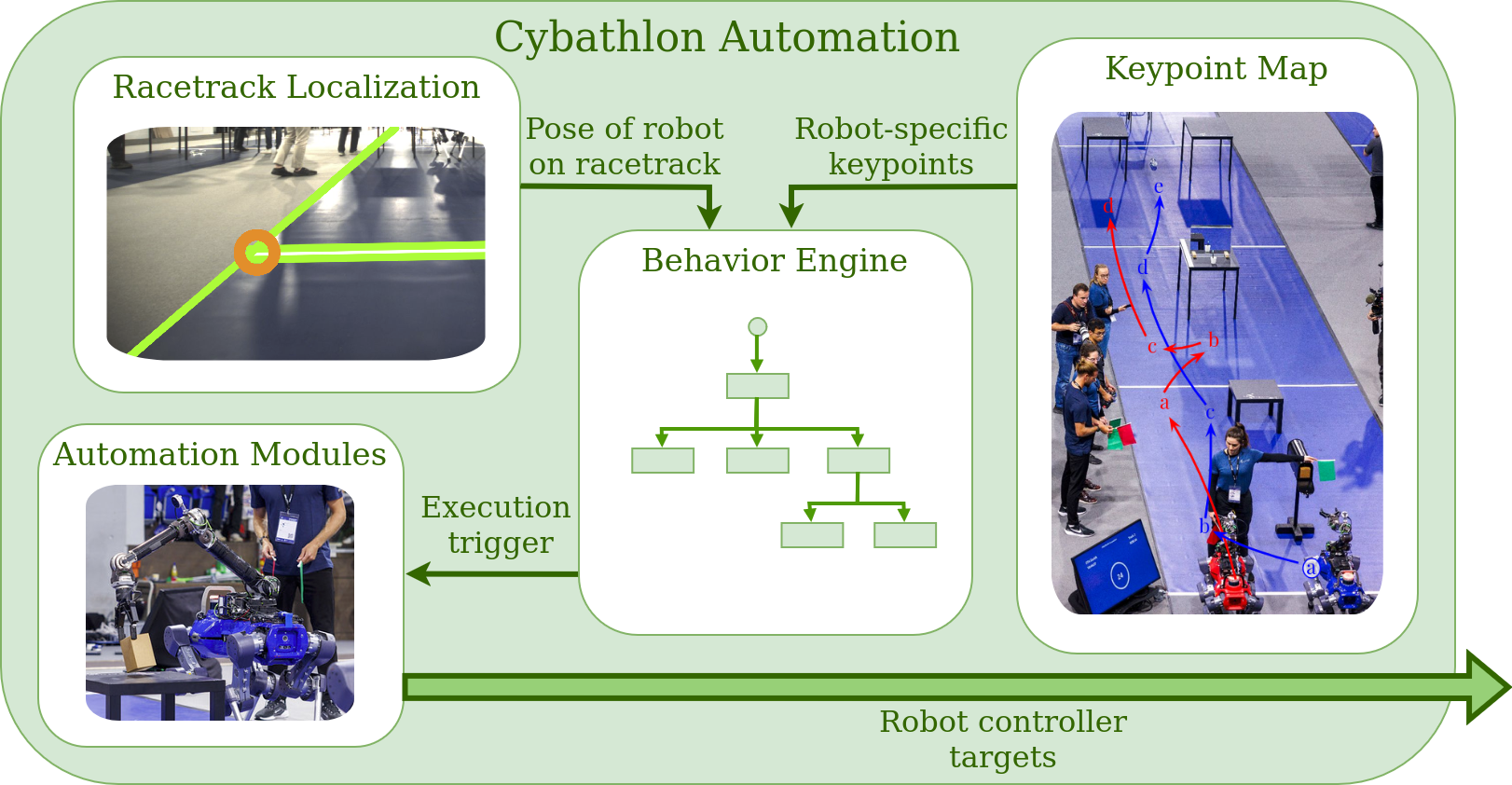}
    \caption{\textbf{The robot automation stack proposed for \textit{Cybathlon} 2024.} Each behavior engine receives its robot-specific keypoints, as well as the current pose of the robot on the racetrack. It keeps track of and communicates its robot's progress and safely steers it through its keypoints, activating the automation modules whenever necessary.}
    \label{fig:nav_cybathlon}
\end{figure*}

From our analysis of the results of the pre-competition of Cybathlon in February 2024, we found that the operator spent approximately 25\% of their time repositioning the robot on the Cybathlon course - something the robots can do autonomously with high reliability and speed. Not only does this movement take time, but it is also challenging for the operator to accurately estimate the robot's location from a fixed viewpoint, and moving themselves and the robot simultaneously is prone to errors in the steering of both. 
As the only team in our race category with an independently moving (i.e., not wheelchair-mounted) assistant robot, this challenge was unique to our approach. We developed the proposed automation stack, visualized in Figure~\ref{fig:nav_cybathlon} and described in this section, to enable the robot to move between tasks autonomously. 
Additionally, this stack leverages the independent base by completing simple, easy-to-automate tasks without operator input.

\emph{Racetrack Localization:}
To autonomously move the robot from task to task along the racetrack, we must first be able to precisely localize it within the racetrack. However, the task of localization on the racetrack is non-trivial: Firstly, the environment is sparse of static features (such as a distinct room layout), as the competition takes place in a massive arena, and only the racetrack obstacles themselves are proximal. Second, the robot is constantly surrounded by many moving agents, including the operator, the robot spotters, the coach, the referee and their assistants, and the camera crew: all of which add significant noise to the environment scans. These factors make conventional \acrfull{slam} methods unreliable. To overcome this, we leverage the strict regularity of the racetrack ground segments and propose a custom localization solution that relies solely on the tape markings that delimit the racetrack sections. More specifically, we aim to detect the corners of the racetrack segments where two tapes intersect perpendicularly. To find these corners, we first perform edge detection using TEED~\cite{soria2023tiny}, which is a minimally sized \acrfull{cnn}. On the resulting edge image, we perform line detection using Hough Transforms, and improve robustness by applying several filtering heuristics, such as requiring the tape lines to have a specific width and to be perpendicular. Between corner detections, the robot uses its own state estimation, which fuses leg odometry and LiDAR and is susceptible to estimation drift. Each new corner observation is then used to correct this drift via a filtered fusion update that integrates previous observations and state estimates, yielding a lightweight and robust localization method designed specifically for the unique challenges of the racetrack environment.



\emph{The Two-Robot Solution:}
Building on the previously described localization method, the robot can autonomously navigate along the racetrack to pre-position itself at upcoming task locations. However, automatic pre-positioning alone does not fully address the time lost during task-to-task transfers, as the operator can still move faster than the robot. To overcome this limitation, we propose a two-robot strategy: while one robot assists the operator with the current task, the other robot moves ahead to prepare for the next task. Once the operator completes the first task, the assisting robot can proceed to the next station, and so on. This approach allows the operator to progress from task to task with minimal idle time waiting for robot repositioning.

\begin{figure*}[h!]
    \centering
    \includegraphics[width=0.95\textwidth]{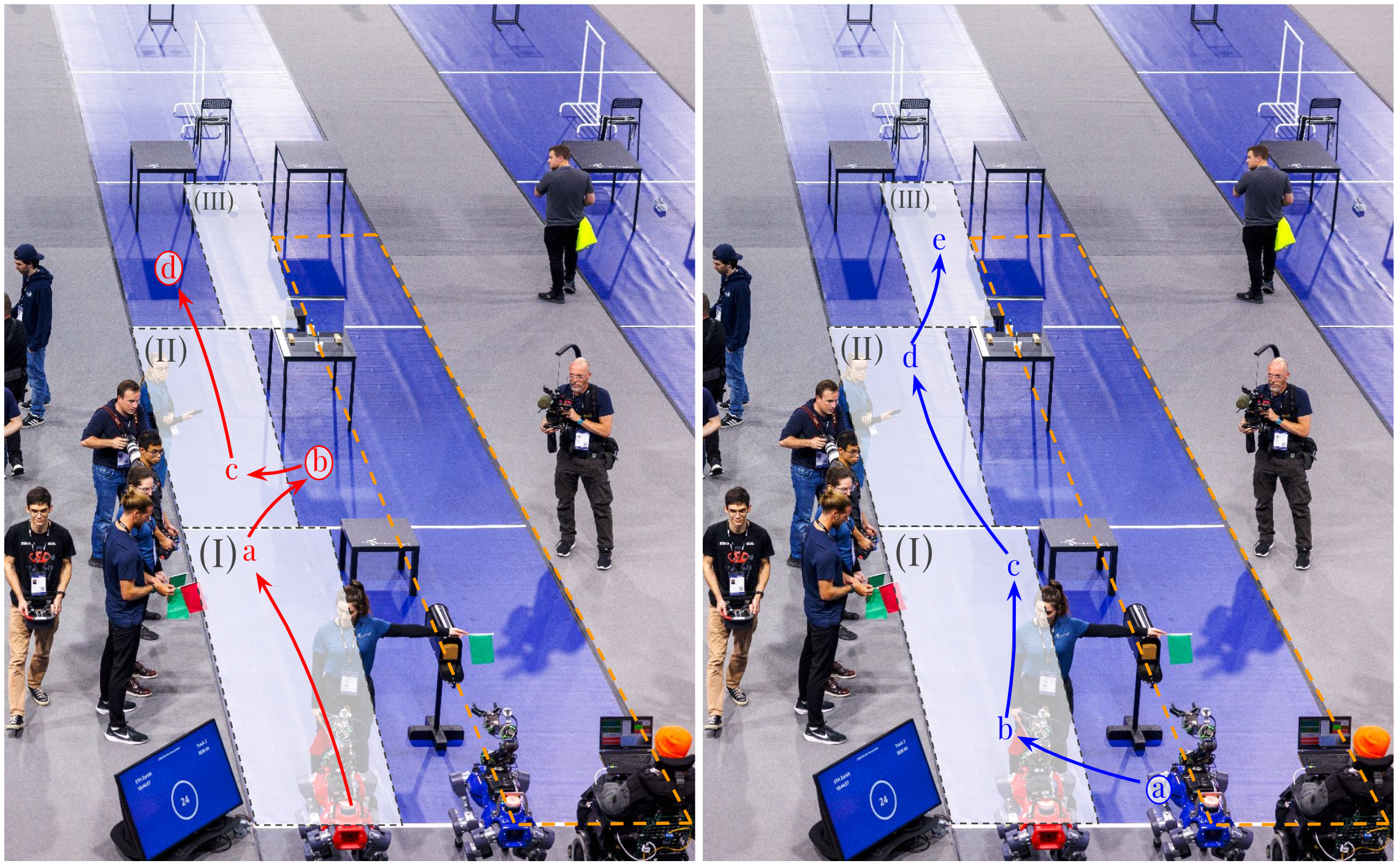}
    \caption{\textbf{The key points for the tasks (I), (II), and the first point of task (III).} Both robots, depicted in red and blue, share the same area definitions, namely the traversal area (indicated in white with grey borders) and the operator area (indicated with an orange border). On the left, the key points for the red robot are visualized, on the right, those of the blue robot. 
    To avoid robot collisions, each robot monitors the other's progress and does not enter a new traversal area if the other robot is already occupying it. For example, since both red point (a) and blue point (b) are in the traversal zone of task (I), if the red robot has not yet reached red point (a) when the blue robot finishes the task at blue point (a), it will not continue to blue point (b) until the red robot has passed red point (a).}
    \label{fig:keypoint_map}
\end{figure*}

\emph{The Keypoint Map:}
To ensure accurate positioning and safe navigation along the track, we define a map of keypoints that guides the robots and prevents collisions with each other and the environment. This map is only made possible by the organizers providing a detailed map of the competition setup. Conceptually, the map can be understood as a connectivity graph of the racetrack, where each task location is linked by collision-free paths through select intermediate points. Figure~\ref{fig:keypoint_map} illustrates the robot-specific keypoints and their connections for tasks one through three. Each key point contains not only the target base position but also a task-specific desired arm configuration and a binary red-light/green-light flag. This last item lets us tackle the central problem arising from this solution: we must coordinate the two robots such that a collision between them becomes impossible. To achieve this, we split each task quadrant (3 meters wide, 5 meters long) into three non-uniform sections. The first is occupied by the operator, giving them enough space to maneuver without fear of the robot colliding with them. The second is occupied by the "working" robot, i.e., the robot that the operator is using to complete the current task. The final area is the traversal area, which the other robot can use to move past both the operator and working robot to the next task. Only one robot can occupy the traversal area at a time. When they then exit the traversal area, they give the green light to the other robot, signaling that it is now safe for them to move. Vice versa, when a robot enters the traversal area, a red light is signaled, blocking the other robot from advancing into the same area. With this strategy, we can ensure that both robots and the operator maintain sufficient distance from one another to avoid collisions, without over-restricting each other.

\emph{The Behavior Engine:}
To ensure correct sequencing of autonomous actions, each robot runs its own \textit{Behavior Tree}, a form of state machine that controls the sequence of autonomous actions the robot undertakes. Initially, the tree receives the list of robot-specific keypoints from the map and constantly monitors its progress through them. The tree directly communicates with the operator via a \acrshort{gui}, signaling when the robots are ready for teleoperation and enabling progression when the operator signals the task as complete. Although each robot has its own independent tree, they are constantly communicating with each other, enforcing the red-light/green-light system to avoid collisions. 

\emph{The Automation Modules:}
In addition to sending the robot along the keypoints of the map, the behavior engine can call a series of modules automating some pick or place motions. This allows us to fully leverage the independence of a mobile system, completing tasks without operator oversight. Primarily, some keypoints act as drop-off locations, indicating that the robot must open its gripper after reaching the target arm configuration. This allows for a two-stage strategy for most pickup tasks: in the first stage, the challenging grasps, such as that of a toothbrush, are teleoperated by the human. After this, and potentially following an interaction between the object and the operator, the robot autonomously returns the object to its designated place, allowing the operator to proceed to the next task. We automate two pickups using task-tailored strategies: the \textit{Apple} pickup location is identified via a color-masking filter tuned to the apple's specific color. Once it is located in the image, we can use the depth values at the corresponding pixels to obtain the 3D position of the apple. The second automated pickup is that of the spices in the \textit{Spice Up} task. We localize the shelf pose with a FoundationPose~\cite{wen2024foundationpose} model, and query the operator for the requested spice via the \acrshort{gui}. Since the spices are at predefined positions within the self, we can easily lift the requested spice and place it on top of the shelf.

Together, these components form an autonomy framework that enables efficient and coordinated operation of multiple robots within the racetrack environment. Precise localization allows each robot to autonomously pre-position for upcoming tasks, while the two-robot strategy minimizes the idle time of the operator. This is further enhanced by leveraging the system's independent nature, enabling it to perform autonomous pick-and-place tasks.

\subsection{At-Home Case Study}
While competition performance provides valuable benchmarks, real-world environments present unique challenges that cannot always be captured in controlled settings. To understand how our robotic system performs in practical, everyday situations, we conducted a case study in an at-home environment. We aimed to evaluate the system's capabilities and limitations in assisting with everyday tasks that the operator cannot perform on their own, as well as the mental load placed on them during system operation. 
The tasks are split into two categories: (I) mixed-method tasks that include the use of the autonomous pre-positioning system and (II) teleoperation-only tasks intended to demonstrate the precision and dexterity our proposed system can reach. These tasks are as follows (I): fetch a snack (apple), fetch a drink (water bottle), pick up and throw away trash (a tissue), microwave food in Tupperware, manipulate the light switch to turn on the lights, (II) pour water from a carafe into a glass and insert a straw, and plug in the operator's wheelchair to charge. The tasks were chosen either by the authors to compare with other works or to demonstrate the system's dexterity (I), or by the operator themself as examples of tasks they would most enjoy assistance with (II).
The tasks were performed consecutively three times; after each run, the participant completed a quick survey to assess the mental load of the system's operation. The survey was completed verbally; the participant was asked to provide their ratings by the corresponding author. The questions asked of the participant are mainly based on NASA TLX~\cite{tlx} and are shown in Table~\ref{tab:questions}, with the grading scale explained in Table~\ref{tab:question_correspondence}. 
The majority of the tasks were not completed zero-shot; for each of them, the operator was given a brief amount of time to familiarize himself with the objects and how the robot could manipulate them. The tasks completed in a zero-shot fasion were \textit{fetch a snack} and \textit{fetch a drink}.

\begin{table}[h]
\caption{\textbf{Questions posed to the operator after completing each test run with the system.} The left column contains the title of the task, as used in the text and the following tables. The right column contains the literal question asked of the operator.}\label{tab:questions}
\begin{tabular}{@{}ll@{}}
\toprule
Question Title & Verbal Question\\
\midrule
Mental Demand & "How mentally demanding was the task?" \\
Temporal Demand & "How hurried or rushed was the pace of the task?"  \\
Performance & "How successful were you in accomplishing what you were asked to do?" \\
Effort & "How hard did you have to work to accomplish your level of performance?" \\
Frustration & "How insecure, discouraged, irritated, stressed, and/or annoyed were you?"  \\
Ease of Use & "How easy did it feel to complete the task?" \\
Error Amount & "How many errors do you feel you made?" \\
Error recovery & "When you made errors, do you feel you recovered successfully?"\\
Subjective Time & "Do you feel the time you took to complete the task was reasonable?" \\
Preference & "Comparing human and robot assistance, which would you prefer?" \\
\botrule
\end{tabular}
\end{table}

\begin{table}[h]
\caption{\textbf{Grade correspondences for the questions posed to the operator}, presented in Table~\ref{tab:questions}. All questions are answered on a scale of 1-7, with 1 being the best and 7 the worst grade.}\label{tab:question_correspondence}
\begin{tabular}{@{}lll@{}}
\toprule
Question Title &  1 corresponds to: & 7 corresponds to: \\
\midrule
Mental Demand  & Not demanding at all & Extremely demanding \\
Temporal Demand  & Not rushed at all & Extremely rushed \\
Performance  & Fully successful & Complete Failure \\
Effort  & Not hard at all & Extremely hard \\
Frustration  & Not bothered at all & Extremely frustrated \\
Ease of Use  & Extremely easy & Extremely hard \\
Error Amount  & No errors whatsoever & Many errors \\
Error recovery  & Recovered easily or made no errors & Failed to recover \\
Subjective Time  & The time was very reasonable & The time was unreasonably long \\
Preference  & Strongly prefer the robot & Strongly prefer the human \\
\botrule
\end{tabular}
\end{table}


\subsubsection{At-Home Automation}
\begin{figure}[h!]
    \centering
    \includegraphics[width=\textwidth]{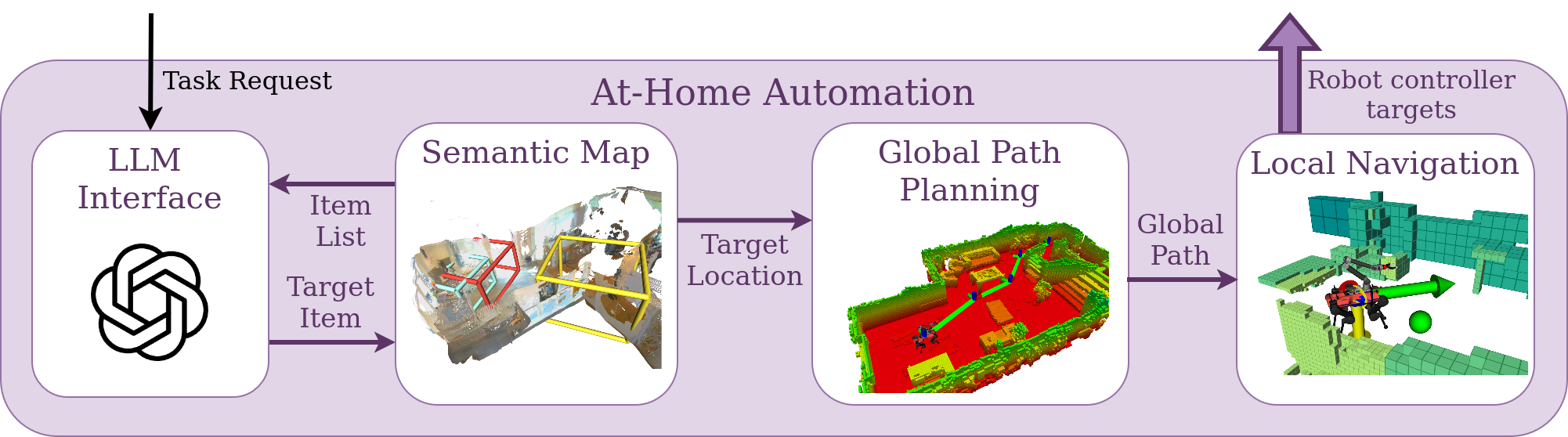}
    \caption{\textbf{Overview of the autonomous mapping and navigation stack} used in the at-home experiments. The input to the stack comes directly from the voice interface as a task request. The \acrshort{llm} interface compares the request to the list of items found in the semantic map and chooses a target. The target location is then passed to a global planner. The planner computes a collision-free path, which is subsequently tracked by the local navigation module.}
    \label{fig:waverider_and_tagmap}
\end{figure}

The \textit{At-Home Automation} stack is an adaptation of the previously described \textit{Cybathlon Automation} stack, tailored to address the challenges of an at-home environment. Naively reproducing the \textit{Cybathlon Automation} stack would include manually tagging every object one might want to interact with at home and creating a connectivity graph of the empty space, which would be very cumbersome. This is further compounded by the fact that any modification to the environment would require repeating this work. To address this, we propose a stack that generates a semantic map~\cite{tagmap} of the full environment, after which the robot can navigate to select objects within the home in a collision-free manner. 
An overview of the system is shown in Figure~\ref{fig:waverider_and_tagmap} and described in detail in this section. 

\begin{figure}[h!]
    \centering
    \includegraphics[width=12cm]{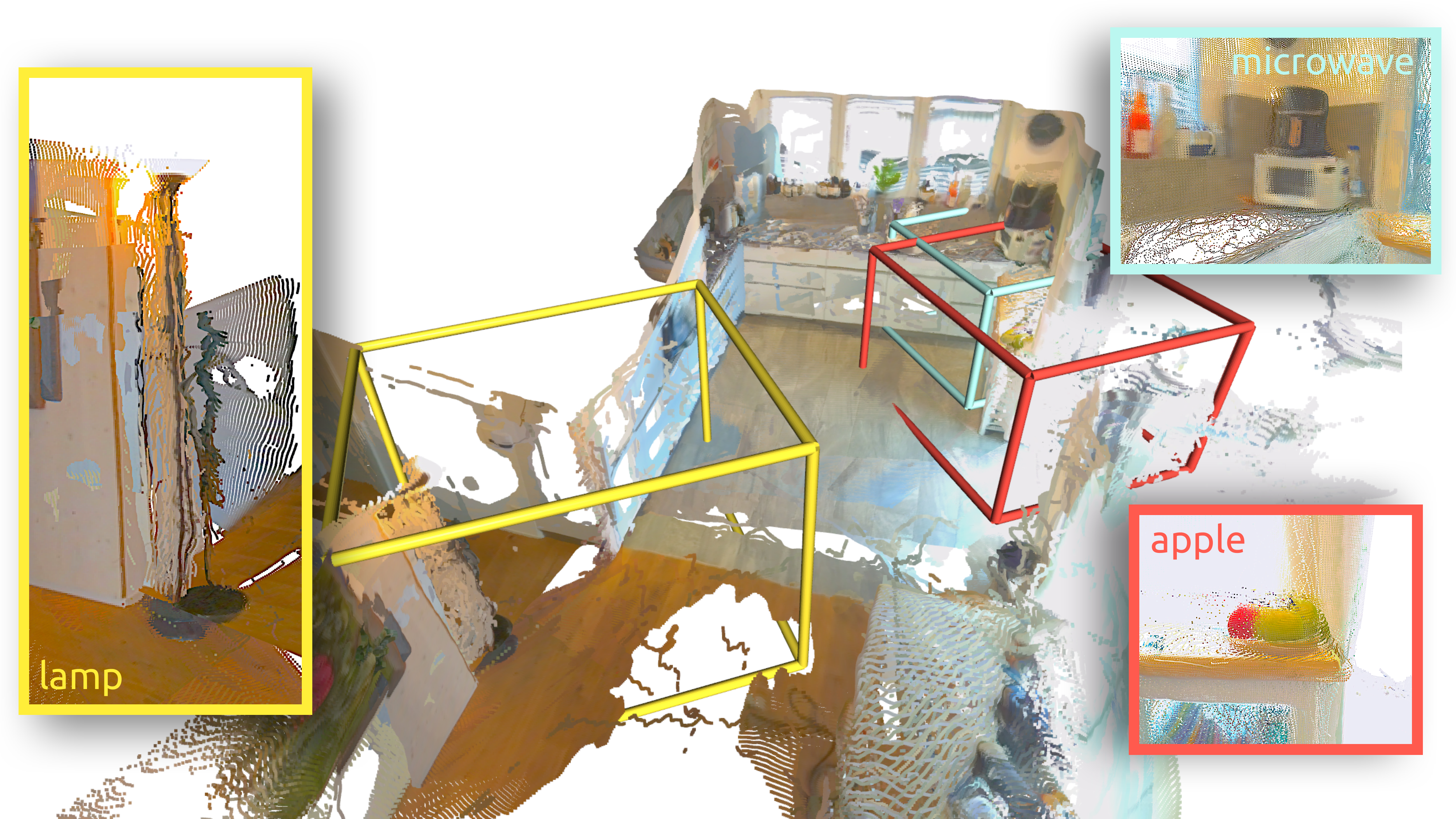}
    \caption{\textbf{The TagMap bounding boxes of the lamp, microwave, and apple}, overlaid over a rendering of the at-home environment used in the study. TagMap lets us localize a large variety of objects within any given scene, providing many candidates for the \acrshort{llm} interface to choose from. This maximizes the likelihood that the interface may find an appropriate match for any given operator request.}
    \label{fig:tagmap_apt}
\end{figure}

The first step of the stack is a speech-to-text interface that processes the participant's voice input, using Whisper~\cite{whisper} for transcription. The raw voice input is not recorded; the interface saves a sentence to be passed along only when the passphrase is detected. As a passphrase, we chose the robot's name, \textit{Donkey}. This command sentence should contain a task request that can range from very specific demands, such as “fetch me an apple” to vague requests such as “I’m feeling like a healthy snack”. The exact wording is open for the participant to choose at their leisure. The full sentence is recorded and passed along to an \acrfull{llm} interface, connected to ChatGPT~\cite{chatgpt} via an API connection. The \acrshort{llm} additionally receives a list of all items found in the environment by the semantic map. Using these two inputs, along with a basic description of the system, the \acrshort{llm} is prompted to identify the most appropriate object to fulfill the participant's request and returns the name of the item it selects. In case no appropriate match can be found, the interface returns failure to the user. This object name is passed back to the semantic map, which contains coarse localizations of all objects found in the environment by an interchangeable image tagging model, in this case \textit{RecognizeAnything}~\cite{ram}. 
Given the name of the target item, TagMap returns its global location in the current map, an example of which is shown in Figure~\ref{fig:tagmap_apt}, to which a global collision-free path is planned using a probabilistic roadmap~\cite{prm} based method on top of the geometric map of the environment~\cite{reijgwart2023efficient}. 
To ensure safety in the actual path-following, the way points of the global path are dispatched to a local navigation system, a custom adaptation of Waverider~\cite{waverider} described in previous work~\cite{scheidemann2025obstacle}, and tracked in a closed-loop fashion. 

The combination of \acrshort{llm} interface, semantic map, and global and local navigation modules allows us to specify tasks in natural language, from which the system can identify and locate an appropriate target object, which can then be navigated to while avoiding collisions with the environment. As soon as the target is reached, control is returned to the operator, who can then manually steer the robot's interaction with the environment to complete the task.


\section{Results}\label{sec:results}

We present two avenues for evaluating the proposed system: controlled benchmarking and real-world testing. While competitions such as \textit{Cybathlon} provide a standardized and high-pressure environment to assess technical robustness, speed, and human–robot coordination, they represent only one facet of our system's capabilities. To gain a more holistic understanding, we combined results from the competition setting with an evaluation in a realistic home environment. This chapter first presents the system’s development, training, and results achieved during \textit{Cybathlon} 2024, before examining its effectiveness and usability in assisting in everyday tasks in a real-world setting.

\subsection{At Cybathlon 2024}
As the system is partially teleoperated, the operator's proficiency is integral to its performance. The amount of training time needed depends largely on how intuitive the system's control scheme is~\cite{10.1097}. During training, the operator practiced executing the competition tasks on a mock race track and provided feedback on refining the control scheme and the \acrshort{gui} visualization of the system. 
Our operator, Samuel Kunz, trained for roughly 40 hours over eleven sessions to reach the level of proficiency exhibited during the final \textit{Cybathlon} competition. This included precision dexterous manipulation, handling both robots on the track at all times, and keeping track of their automation module progression. Competition videos are available on the official Cybathlon website\footnote{\url{https://cybathlon.com/en/cybathlon-2024/results?event=5267&discipline=5153}}.

\emph{\textbf{Strategy:}}
Prior to the competition runs, we had decided to cross the task field while bypassing the \textit{Door} task. While our system was physically capable of performing the task, the full sequence (teleoperating the door opening, navigating the operator and both robots through, and finally closing the door) took approximately two minutes during practice runs. Attempting it during the full race would have introduced a significant risk of exceeding the 10-minute time limit, potentially losing us more points on the subsequent tasks. The door opening task was found to be the most cumbersome for our robot embodiment, as the operator and both robots must navigate through the door, whereas wheelchair-mounted systems only require the pilot to move themselves through the obstacle. Consequently, we focused on completing the remaining nine tasks, shown in Figure~\ref{fig:track}.

\emph{\textbf{The Qualifiers:}}
The preliminary competition consisted of two runs, and for each team, the run with the best score (best time on equal scores between runs) was counted as the final result. 
On our first run, we completed eight of the ten tasks in 7 minutes 49 seconds. The two unsolved tasks were \textit{Spice Up} and \textit{Door}. For the \textit{Spice Up} task, an error in the automated pick-up procedure caused a collision of one of the spice bottles with the shelf, prompting immediate failure of the task.
On the second run, a user error of the operator (moving to manual control before the robot had fully arrived at its target keypoint) caused a total failure of the behavior engine, forcing the operator to move to full manual control of both robots. The time loss due to this was palpable on the last task, \textit{Dishwasher}, where we were interrupted before completion because the 10-minute time limit had been reached. Additionally, \textit{Toothbrush} was judged incomplete due to the brush falling off the table after it was returned, and \textit{Spice Up} had to be skipped due to the termination of the behavior engine. In total, we successfully solved six tasks during the second run. 

\emph{\textbf{The Final:}}
The final round of the competition consisted of a single round on the racetrack. As in our best qualifying round, we successfully solved eight of the ten tasks. However, on this run \textit{Spice Up} was completed successfully, but a small teleoperation error disqualified us from receiving points on the \textit{Mailbox} task. As in the qualifying rounds, we skipped the \textit{Door} task entirely. In total, we cleared the track in 8 minutes and 12 seconds. 

\emph{\textbf{Time Management:}}
Extending the automation stack to include autonomous pre-positioning and simple object pickup reduced our overall time on the \textit{February 2024 Challenge} tasks from 6 minutes 34 seconds to 4 minutes 52 seconds. Of that time, the operator waited only approximately 21 seconds for the robot to position itself, equivalent to 7\% of task time. When comparing with the 25\% of task time spent on moving at the \textit{February 2024 Challenge}, the advantages of automatic pre-positioning, even without considering the benefit of alleviating the mental load, are stark.

\begin{table}[h]
    \centering
    \begin{tabular}{|c|cccccccccc|}
        \hline
        Task & (I) & (II) & (III) & (IV) & (V) & (VI) & (VII) & (VIII) & (IX) & (X) \\
        \hline
        EDAN  & 0:52 & 0:40 & \textbf{0:41} & \textbf{0:54} & 0:47 & \textbf{0:18} & 1:24 & \textbf{1:37} & 0:32 & DNF \\
        SmartArM  & \textbf{0:34} & 0:51 & 0:53 & DNF & 0:53 & 0:19 & 1:54 & 2:06 & 0:28 & 1:45 \\
        RSL (Official)  & DNF & 1:08 & 0:42 & 0:42 & 1:31 & 0:19 & 1:40 & DNF & 0:59 & 1:11\\
        RSL (Custom)  & DNF & \textbf{0:32} & 0:57 & 1:10 & \textbf{0:37} & 1:45 & \textbf{1:18} & DNF & \textbf{0:22} & 1:54\\
        BFH-FAIR & 0:48 & 0:46 & 0:48 & 0:55 & 1:04 & DNF & DNF & DNF & 1:03 & \textbf{1:26} \\
        \hline
    \end{tabular}
    \caption{\textbf{Task completion times from the finals of the Assistance Robot Race at \textit{Cybathlon} 2024 for all teams.} Detailed task descriptions are given in Section~\ref{sec:comp_desc}. Tasks marked with \acrfull{dnf} were incomplete and therefore not timed. Under the official rules, a task is complete and timing stops when the last agent (operator or robot) leaves the task field; these values are reported in RSL (Official). Due to the \emph{Two-Robot Solution}, our operator often advanced to the next task while one robot finished the current one autonomously, making task boundaries ambiguous. We therefore also report RSL (Custom), where task progression is defined as the moment the operator marked a task complete in the \emph{Behavior Engine} and shifted attention to the next task. We believe this better reflects the time the operator actively spent per task. The lowest completion time per task is highlighted in \textbf{bold}, without consideration of RSL (Official).}
    \label{tab:task_completion_times}
\end{table}
\vspace{-20pt}

\emph{\textbf{Comparison to Other Teams:}}
In the qualifiers, we placed fourth (fifth if the two BFH-Fair operators are counted separately). This narrowly secured us a spot in the final round, which was open to the top four teams. There, we earned the third place, positioned behind the teams of EDAN and SmartArM, which had both solved nine tasks each, and ahead of BFH-FAIR, which had completed seven tasks. Of all competing teams, we presented the only independent mobile robotic assistant, as well as the only multi-robot solution. The operators across teams had differing levels of residual mobility, which influenced each team’s choice of user interface. For example, EDAN relied on a SpaceMouse and a tablet interface~\cite{doi:10.1126/scirobotics.aeb6725}, while the SmartArM operator directly steered their wheelchair using the chair’s wheels and their robotic arm via a tablet.
As shown in Table~\ref{tab:task_completion_times}, under the adjusted timing method, we achieved the fastest completion times in (II) Toothbrush, (V) Eating, (VII) Spice Up, and (IX) Touchscreen. These results highlight the benefit of autonomous task execution. In (II), the robot autonomously returned the toothbrush to the table while the pilot moved on. In (V), it picked up the apple before the pilot arrived and put it back down after they had left. In (VII), the robot completed the full spice pick-and-place sequence autonomously, requiring only high-level item selection from the operator. In (IX), the robot accurately prepositioned itself in front of the tablet, ready for the operator to click on the correct menu item before they had even entered the task field.
The complications of bringing two independent systems onto the narrow racetrack are highlighted with tasks (VI) Crowd and (VIII) Door. Navigating not only the pilot but also two robotic systems through the crowd obstacle increased the execution time to more than five times that of the fastest team. However, under the original timing metric (defined as the time needed for the last agent, the second robot, to move through the field), our performance is comparable to that of the other teams. This additional coordination effort also prevented us from attempting the door task, as discussed in the \emph{Strategy} section. In tasks (III) Pick-Up, (IV) Scarf, and (X) Dishwasher, where task parallelization could not be leveraged as effectively, our execution times were slightly higher than those of other teams.


\subsection{At Home}
In the at-home setting, we evaluate over two classes of tasks: those that only require teleoperation, and those that additionally leverage the at-home autonomy stack. The teleoperation-only tasks were handpicked by the operator as examples of tasks he would most enjoy assistance with. They are included to further showcase the dexterity of the proposed system and its applicability to tasks unseen during system development. Subsequently, we present the shared autonomy tasks, which aim to additionally highlight the versatility of the at-home autonomy assistance.

\subsubsection{Teleoperation-Only Tasks}

\begin{figure}[h!]
    \centering
    \includegraphics[width=12.5cm]{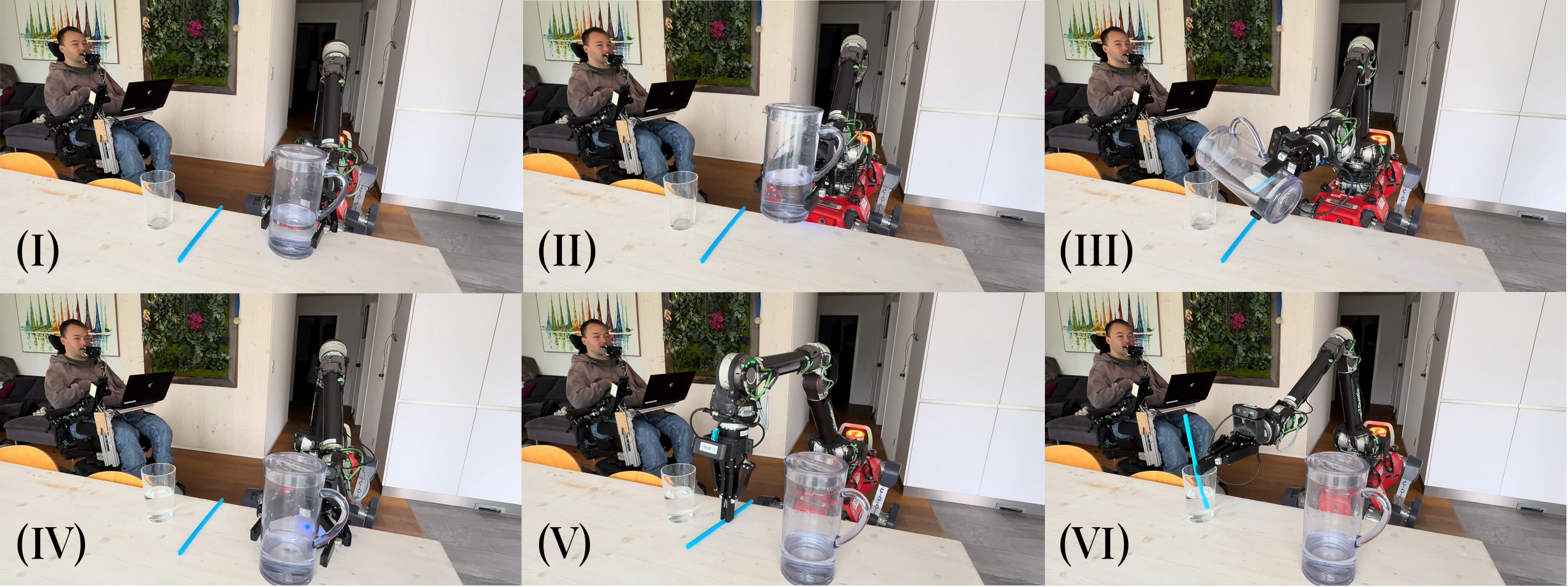}
    \caption{\textbf{Image sequence of the \emph{Pour Water Glass} task}. Description from left to right, top to bottom: (I) The carafe is grabbed. (II) The carafe is lifted. (III) Water is poured from the carafe into the glass. (IV) The carafe is set back down on the table. (V) The straw is picked up from the table. (VI) The straw is inserted into the glass.}
    \label{fig:pour_water_glass}
\end{figure}

\begin{figure}[h]
  \centering
  \includegraphics[width=0.75\linewidth]{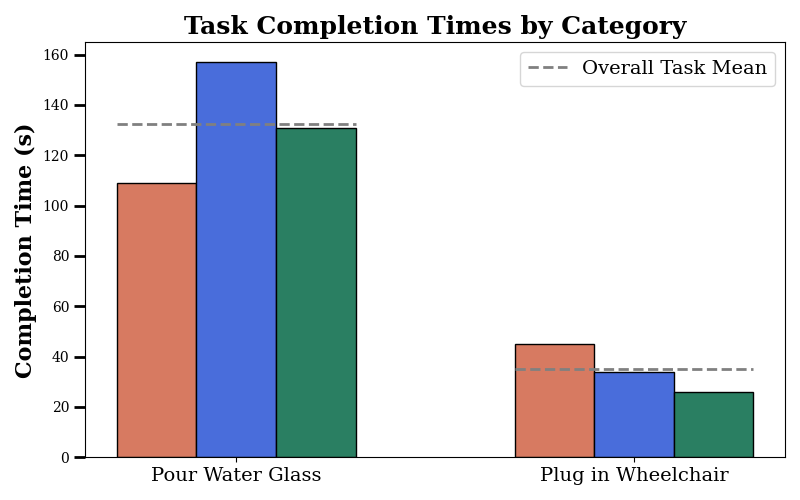}
    \caption{\textbf{Progression in task completion times over multiple runs for the teleoperation-only tasks.} A high variance in completion times can be observed for the \emph{Pour Water Glass} task runs, whereas \emph{Plug in Wheelchair} shows a more consistent decrease in time needed.}
    \label{fig:times_teleop}
\end{figure}

\emph{\textbf{Experimental Results}}
In the \emph{Pour Water Glass} task, the dexterity of the system is visualized in Figure~\ref{fig:pour_water_glass}: the operator effortlessly lifts the heavy carafe, turns it to the precise angle needed to pour out the water without spilling, sets it back down, then grasps the delicate straw. The \emph{Plug in Wheelchair} task is a sophisticated peg-in-hole problem, as it requires the operator to pre-align the plug with the socket before inserting it with a moderate amount of force. As shown in Table~\ref{tab:user_data_teleop}, both are tackled very successfully: across the two sets of task runs conducted in the teleoperation-only format, the operator completed all runs without requiring task resets. As such, no failure reason can be reported. The times taken for each task are not directly comparable, as the \emph{Pour Water Glass} task consists of multiple subsequent steps, whereas \emph{Plug in Wheelchair} can be completed in a single step. The variance of task completion times in between runs is visible in Figure~\ref{fig:times_teleop}. We can observe a high variance in completion times for the \emph{Pour Water Glass} task, whereas \emph{Plug in Wheelchair} shows a smoother decrease. This may indicate that tasks of higher complexity, such as \emph{Pour Water Glass}, may take longer for the operator to master.

\begin{table}[h!]
\begin{tabular}{@{}llll@{}}
\toprule
Scale Title & \textbf{Pour Water Glass} & \textbf{Plug in Wheelchair}\\
\midrule
Completion Time (Mean) & 133s & 35s \\
Success Rate & 3/3 & 3/3 \\
\botrule
\end{tabular}
\caption{\textbf{Quantitative results of the teleoperation-only experiments.} Task completion time is reported as the mean recorded value over three runs. For the success rate, any physical intervention is treated as a failure.}\label{tab:user_data_teleop}
\end{table}

\emph{\textbf{User Study}}
As shown in Table~\ref{tab:user_response_teleop}, the operator reports a fairly high mental demand while completing the presented tasks. This is to be expected, as both of the tasks require a high degree of dexterity. This is particularly visible on the \emph{Effort} response, which is fairly high for the \emph{Plug in Wheelchair} task, as this requires a high degree of spatial reasoning. The operator's confidence and calm, gained in part through training hours in preparation for Cybathlon 2024, are evident in the low values of the \emph{Performance} and \emph{Frustration} responses. The combination of this with the intuitiveness of the proposed system results in perfect scores on \emph{Error Amount} and near-perfect scores on \emph{Ease of Use}. It is interesting to note that although the operator found the \emph{Plug in Wheelchair} task harder to complete, they were very satisfied with the time it took them. Inversely, the task perceived as easier, \emph{Pour Water Glass}, left them less satisfied with their speed. When compared with human assistance, the trade-off between greater independence and additional labor seems to even out, as the operator reports no preference for either.

\begin{table}[h!]
    \begin{minipage}{.8\linewidth}
        \begin{tabular}{@{}llll@{}}
            \toprule
            Scale Title & \makecell{\textbf{Pour Water} \\ \textbf{Glass}} & \makecell{\textbf{Plug In} \\ \textbf{Wheelchair}}\\
            \midrule
            Mental Demand & \cellcolor{BrickRed!55}4 & \cellcolor{BrickRed!55}4 \\
            Temporal Demand & \cellcolor{BrickRed!25}2 & \cellcolor{BrickRed!25}2 \\
            Performance & \cellcolor{BrickRed!10}1 & \cellcolor{BrickRed!10}1 \\
            Effort & \cellcolor{BrickRed!40}3 & \cellcolor{BrickRed!70}5 \\
            Frustration & \cellcolor{BrickRed!10}1 & \cellcolor{BrickRed!25}2 \\
            Ease of Use & \cellcolor{BrickRed!25}2 & \cellcolor{BrickRed!10}1 \\
            Error Amount & \cellcolor{BrickRed!10}1 & \cellcolor{BrickRed!10}1 \\
            Error recovery & \cellcolor{BrickRed!10}1 & \cellcolor{BrickRed!10}1 \\
            Subjective Time & \cellcolor{BrickRed!55}4 & \cellcolor{BrickRed!10}1 \\
            Preference & \cellcolor{BrickRed!55}4 & \cellcolor{BrickRed!55}4 \\
            \botrule
            \end{tabular}
    \end{minipage}
    \qquad
    \begin{minipage}{.2\linewidth}
        \begin{tabular}{@{}l@{}}
            \toprule
            Color Scale\\
            \midrule
            \cellcolor{BrickRed!10}1 \\
            \cellcolor{BrickRed!25}2 \\
            \cellcolor{BrickRed!40}3 \\
            \cellcolor{BrickRed!55}4 \\
            \cellcolor{BrickRed!70}5 \\
            \cellcolor{BrickRed!85}6 \\
            \cellcolor{BrickRed!100}7 \\
            \botrule
            \end{tabular}
    \end{minipage} 
    \caption{\textbf{The responses to the questions detailed in Table~\ref{tab:questions}, after each run of the corresponding task of the \textit{teleoperation-only} category.} Median value over three runs per task. The scale ranges from 1-7, with 1 being the best grade over all categories, defined as described in Table~\ref{tab:question_correspondence}.}\label{tab:user_response_teleop}
    \vspace{-30pt}
\end{table}

\subsubsection{Shared Autonomy Tasks}
\begin{figure}[h!]
    \centering
    \includegraphics[width=12.5cm,trim={0 0 0 0.2cm},clip]{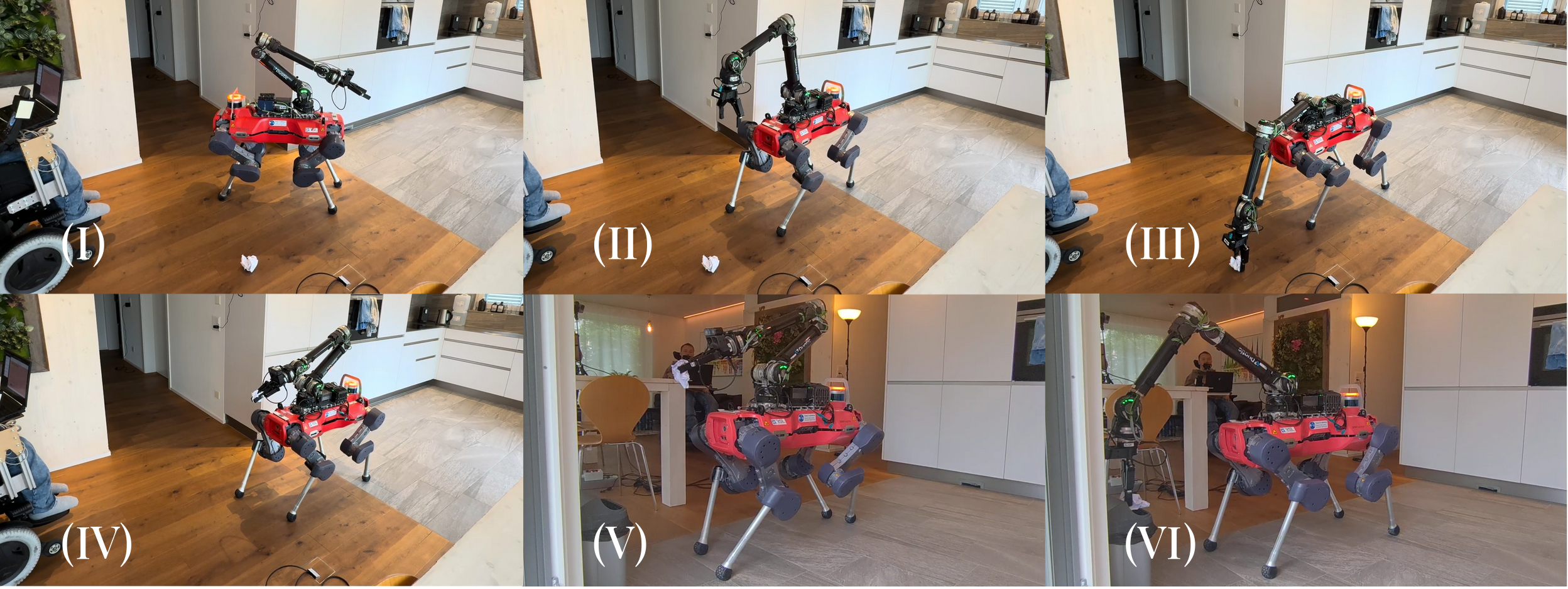}
    \caption{\textbf{Image sequence of the \emph{Throw Trash} task.} Description from left to right, top to bottom: (I) The operator takes control of the robot. (II) The operator positions the robot so that it faces the tissue. (III) The operator grabs the tissue on the floor. (IV) The operator asks the robot, "Go to the trash can". (V) The robot finds "bin" in the TagMap and approaches it. (VI) The operator throws the tissue into the trash bin.}
    \label{fig:trash}
    \vspace{-10pt}
\end{figure}

\begin{figure}[h!]
    \centering
    \includegraphics[width=12.5cm]{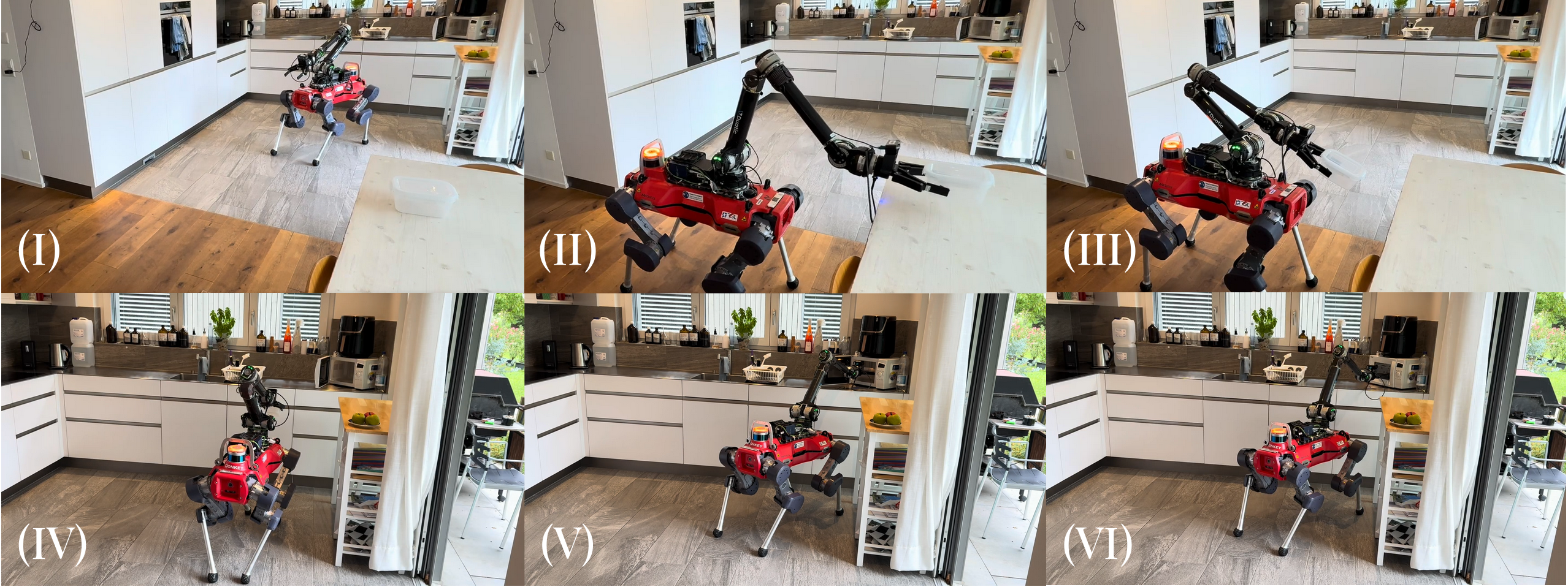}
    \caption{\textbf{Image sequence of the \emph{Warm Up Food} task.} Description from left to right, top to bottom: (I) The robot starts from an initial position in the middle of the room. (II) The operator picks up the Tupperware. (III) The operator asks the robot, "I want to warm up my food". (IV) The robot finds "microwave" in the TagMap and approaches it. (V) The operator puts the Tupperware in the microwave. (VI) The operator closes the microwave door.}
    \label{fig:warm_up_food}
    \vspace{-10pt}
\end{figure}

\begin{figure}[h]
    \centering
    \includegraphics[width=13cm]{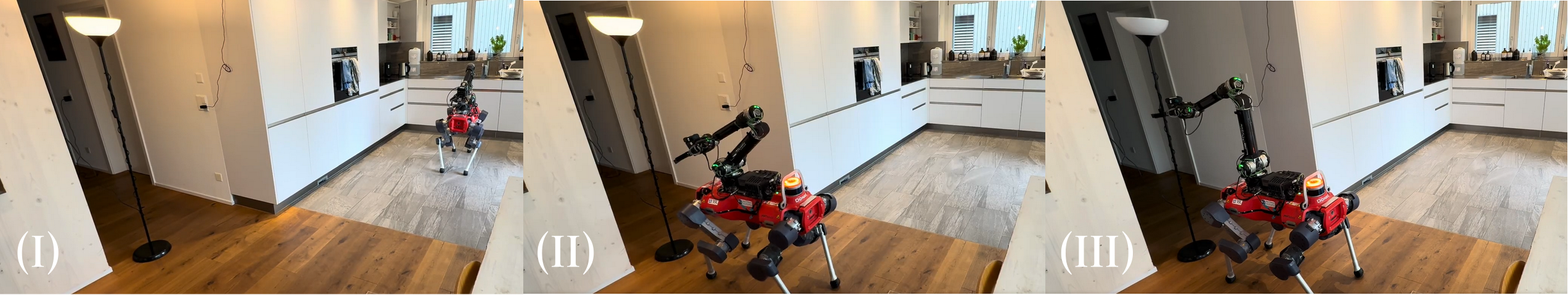}
    \caption{\textbf{Image sequence of the \emph{Lights} task.} Description from left to right: (I) The operator asks the robot, "Turn off the lights". (II) The robot finds "lamp" in the TagMap and approaches it. (III) The operator presses the light switch.}
    \label{fig:lights}
    \vspace{-10pt}
\end{figure}

\begin{figure}[h!]
  \centering
  \includegraphics[width=0.95\linewidth]{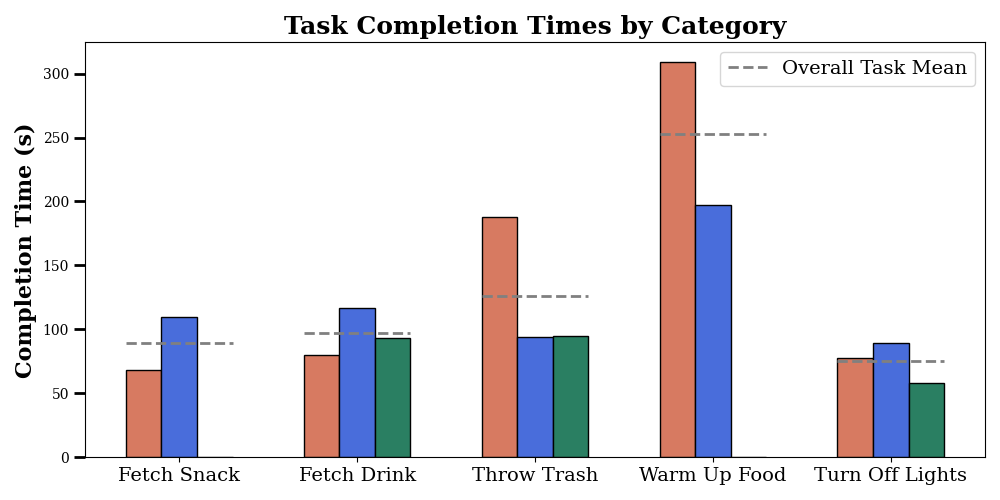}
    \caption{\textbf{Progression in task completion times over multiple runs for the shared autonomy tasks.} A high variance in completion times can be observed for the \emph{Throw Trash} and \emph{Warm Up Food} task runs, which require more precise manipulation than, for example, the \emph{Fetch Drink} task. A missing bar indicates that the run was not successful and therefore was not timed.}
    \label{fig:times_mixed}
\end{figure}

\emph{\textbf{Experimental Results}}
We showcase performance of the system over a variety of tasks, from the more straightforward "find the object, approach the object, interact with the object" tasks (\emph{Fetch Snack}, \emph{Fetch Drink}, \emph{Lights}), to those that require multiple stages of manipulation (\emph{Throw Trash}, \emph{Warm Up Food}). Multiple of these tasks require precision dexterous manipulation, such as picking up the tissue in the \emph{Throw Trash} task (Figure~\ref{fig:trash}), inserting the Tupperware into the microwave and closing it in the \emph{Warm Up Food} task (Figure~\ref{fig:warm_up_food}), and flipping the small switch on the lamp in the \emph{Lights} task (Figure~\ref{fig:lights}).
The varying complexities of the tasks are also evident in their average completion times shown in Table~\ref{tab:user_data_mixed}: the most complex, \emph{Warm Up Food}, takes over twice as long as the simple fetch tasks. Overall, the completed tasks highlight the breadth of tasks the proposed system can tackle, showcasing its application flexibility.
Of the 15 task runs started, 13 were completed successfully, without the operator or externals intervening in the automation or the environment. In one of the 15 runs, the voice interface picked up the words spoken by the operator incorrectly, leading the robot to walk to the wrong object. In the other instance, the target goal calculation failed, requiring external intervention. It is worth noting that, in both cases, although we judged the task as failed, the operator would still have been able to switch from automation to manual control and complete the task via teleoperation. The progression of task completion times is visible in Figure~\ref{fig:times_mixed}. Although the data does not indicate a clear cross-category trend, as for the teleoperation-only tasks, we here again observe a higher variance for tasks with more complex manipulation, such as \emph{Throw Trash} and \emph{Warm Up Food}. 

\begin{table}[h!]
\begin{tabular}{@{}llllll@{}}
\toprule
Scale Title & \makecell{\textbf{Fetch} \\ \textbf{Snack}} & \makecell{\textbf{Fetch} \\ \textbf{Drink}} & \makecell{\textbf{Throw} \\ \textbf{Trash}} & \makecell{\textbf{Warm Up} \\ \textbf{Food}} & \makecell{\textbf{Turn Off} \\ \textbf{Lights}} \\
\midrule
\makecell{Task Completion\\Time (Mean)} & 89s & 97s & 126s & 253s & 75s
\vspace{5pt}\\
\makecell{Success Rate} & 2/3 & 3/3 & 3/3 & 2/3 & 3/3
\vspace{5pt}\\
\makecell{Failure Reason \\ (If Applicable) } & \makecell{Error in the\\goal calculation}  & - & - & \makecell{Failure of the \\ voice interface} & - \\
\botrule
\end{tabular}
\caption{\textbf{Quantitative results of the shared autonomy experiments.} Task completion time is reported as the mean recorded value over three runs. For the success rate, any physical intervention or interruption of the automation is treated as a failure.}\label{tab:user_data_mixed}
\vspace{-20pt}
\end{table}

\emph{\textbf{User Study}}
Overall, from Table~\ref{tab:user_response_mixed}, it is clear that the mixed-method tasks felt less challenging to the operator (lower \emph{Mental Demand} and \emph{Effort} scores) than the teleoperation-only tasks. We attribute this in part to the nature of the mixed-method tasks, which include some less dexterous tasks, namely \emph{Fetch Snack}, \emph{Fetch Drink}, and \emph{Lights}. However, \emph{Throw Trash} and \emph{Warm Up Food} both contain challenging precision manipulation and still score lower than teleoperation-only. From this, we conclude that the automation of the pre-positioning base motion indeed relieves the operator's mental load. However, we see a small increase in \emph{Frustration} and \emph{Ease of Use} categories. During the experiments, we observed that the speech-to-text interface had difficulty correctly interpreting the operator's speech, leading him to repeat his requests multiple times. The operator again gives low scores in \emph{Temporal Demand}, \emph{Performance}, \emph{Error Amount and Recovery}, and \emph{Subjective Time}. Compared to the teleoperation-only tasks, we see a shift towards robot preference in this category, with the average score reducing to 3.2 (compared to the previous 4). This suggests that a higher degree of autonomy makes the system more attractive for everyday use.

\begin{table}[h!]
    \begin{minipage}{.8\linewidth}
        \begin{tabular}{@{}llllll@{}}
        \toprule
        Scale Title & \makecell{\textbf{Fetch} \\ \textbf{Snack}} & \makecell{\textbf{Fetch} \\ \textbf{Drink}} & \makecell{\textbf{Throw} \\ \textbf{Trash}} & \makecell{\textbf{Warm Up} \\ \textbf{Food}} & \makecell{\textbf{Turn Off} \\ \textbf{Lights}} \\
        \midrule
        Mental Demand & \cellcolor{BrickRed!25}2 & \cellcolor{BrickRed!25}2 & \cellcolor{BrickRed!25}2 & \cellcolor{BrickRed!40}3 & \cellcolor{BrickRed!25}2 \\
        Temporal Demand & \cellcolor{BrickRed!25}2 & \cellcolor{BrickRed!10}1 & \cellcolor{BrickRed!10}1 & \cellcolor{BrickRed!40}3 & \cellcolor{BrickRed!25}2 \\
        Performance & \cellcolor{BrickRed!10}1 & \cellcolor{BrickRed!10}1 & \cellcolor{BrickRed!10}1 & \cellcolor{BrickRed!10}1 & \cellcolor{BrickRed!10}1  \\
        Effort & \cellcolor{BrickRed!25}2 & \cellcolor{BrickRed!25}2 & \cellcolor{BrickRed!25}2 & \cellcolor{BrickRed!25}2 & \cellcolor{BrickRed!25}2  \\
        Frustration & \cellcolor{BrickRed!40}3 & \cellcolor{BrickRed!10}1 & \cellcolor{BrickRed!25}2 & \cellcolor{BrickRed!25}2 & \cellcolor{BrickRed!25}2  \\
        Ease of Use & \cellcolor{BrickRed!25}2 & \cellcolor{BrickRed!25}2 & \cellcolor{BrickRed!25}2 & \cellcolor{BrickRed!40}3 & \cellcolor{BrickRed!40}3  \\
        Error Amount & \cellcolor{BrickRed!10}1 & \cellcolor{BrickRed!40}3 & \cellcolor{BrickRed!10}1 & \cellcolor{BrickRed!40}3 & \cellcolor{BrickRed!10}1  \\
        Error recovery & \cellcolor{BrickRed!10}1 & \cellcolor{BrickRed!10}1 & \cellcolor{BrickRed!10}1 & \cellcolor{BrickRed!40}3 & \cellcolor{BrickRed!10}1  \\
        Subjective Time & \cellcolor{BrickRed!10}1 & \cellcolor{BrickRed!25}2 & \cellcolor{BrickRed!25}2 & \cellcolor{BrickRed!25}2 & \cellcolor{BrickRed!25}2  \\
        Preference & \cellcolor{BrickRed!40}3 & \cellcolor{BrickRed!40}3 & \cellcolor{BrickRed!40}3 & \cellcolor{BrickRed!55}4 & \cellcolor{BrickRed!40}3  \\
        \botrule
        \end{tabular}
    \end{minipage}
    \qquad
    \begin{minipage}{.2\linewidth}
        \begin{tabular}{@{}l@{}}
            \toprule
            Color Scale\\
            \midrule
            \cellcolor{BrickRed!10}1 \\
            \cellcolor{BrickRed!25}2 \\
            \cellcolor{BrickRed!40}3 \\
            \cellcolor{BrickRed!55}4 \\
            \cellcolor{BrickRed!70}5 \\
            \cellcolor{BrickRed!85}6 \\
            \cellcolor{BrickRed!100}7 \\
            \botrule
            \end{tabular}
    \end{minipage} 
    \caption{\textbf{The responses to the questions detailed in Table~\ref{tab:questions}, after each run of the corresponding task of the \textit{shared autonomy} category.} Median value over three runs per task. The scale ranges from 1-7, with 1 being the best grade over all categories, defined as described in Table~\ref{tab:question_correspondence}.}
    \label{tab:user_response_mixed}
    \vspace{-30pt}
\end{table}

\subsubsection{Comparing to other State-of-the-Art}

\begin{figure}[h!]
    \centering
    \includegraphics[width=10cm]{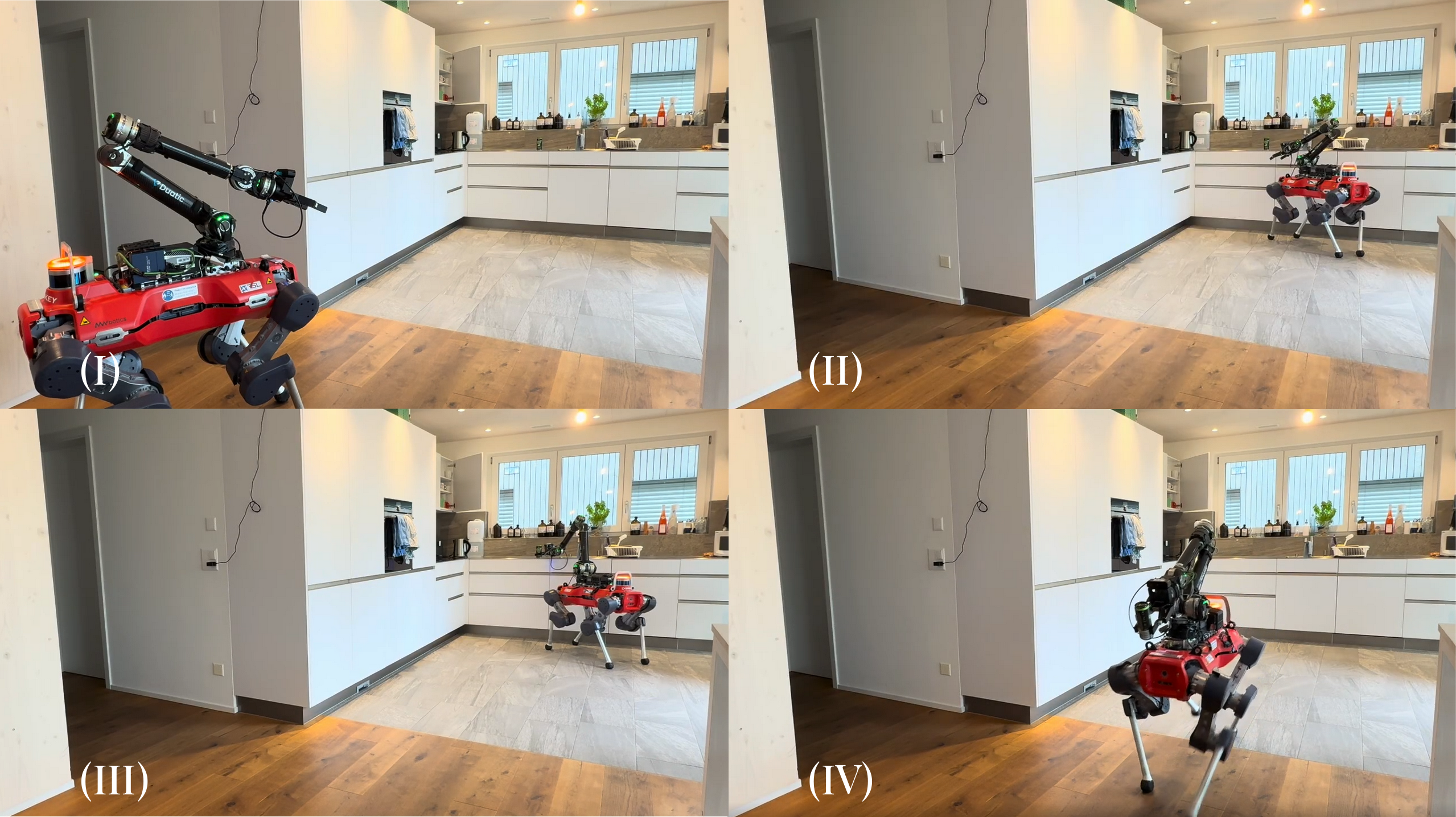}
    \vspace{5pt}
    \caption{\textbf{Image sequence of the \emph{Fetch Drink} task.} Description from left to right, top to bottom: (I) The operator tells the robot, "Get me a soda can". (II) The robot finds "bottle" in the TagMap and approaches it. (III) The operator picks up the can. (IV) The operator steers the robot back to themselves.}
    \label{fig:drink}
\end{figure}

\begin{table}[h!]
\centering
\begin{tabular}{@{}llll@{}}
\toprule
Scale Title & \textbf{HAT} & \textbf{Our System}\\
\midrule
Mental Demand & 2 & 2.2 \\
Temporal Demand & 1 & 1.8 \\
Performance & 1 & 1.0 \\
Effort & 2 & 2.0 \\
Frustration & 1 & 2.0 \\
Ease of Use & 2 & 2.4 \\
Error Amount & 2 & 1.8 \\
Error recovery & 2 & 1.6 \\
Subjective Time & 2 & 1.8 \\
Preference & 3 & 3.2 \\
\botrule
\end{tabular}
\caption{\textbf{Comparison of the responses of our operator and those reported by Padmanabha et al.~\cite{padmanabha2024independence} for their system, HAT.}  The responses of the HAT system are translated to match the scale used by the authors: lower is better. For our system, we report the mean of the results of the mixed-method experiments only.}\label{tab:hri_comp}
\vspace{-15pt}
\end{table}

We compare the results of our user study with those of Padmanabha \textit{et al.}~\cite{padmanabha2024independence}, whose work is elaborated upon in Section~\ref{sec:background}. We note that this is a rough comparison, as neither our work nor theirs presents a benchmark, and we are comparing the subjective experiences of the two expert operators. Additionally, we compare our results only to those reported on the final day of their study, as our operator had prior experience operating the system at the time of the survey. Both systems utilize a form of partial teleoperation and partial automation. The comparison can be seen in Table~\ref{tab:hri_comp}. Furthermore, Padmanabha \textit{et al.} report a task completion time of approximately 3 minutes for their \textit{Fetching Red Bull} task. We aimed to emulate this task with our \textit{Fetch Drink}, visible in Figure~\ref{fig:drink}, and report an average task completion time of 1 minute 37 seconds, as reported in Table~\ref{tab:user_data_mixed}.

From the results in Table~\ref{tab:hri_comp}, it is visible that both systems score comparatively. However, the system proposed in this work enables a wider range of tasks, thanks to the legged base's increased versatility and the arm's larger workspace and payload capacity.

\section{Discussion}\label{sec:discussion}



\subsection{Cybathlon}
At the Cybathlon 2024 competition, ours was the only team employing an independent mobile robotic platform. The results demonstrate that the proposed system performs comparably to state-of-the-art wheelchair-mounted robotic arms, the strategy employed by all other teams. However, our platform’s independence and shared-autonomy framework present significant advantages, enabling the operator to complete tasks successfully and efficiently despite our system relying the least on residual upper-limb mobility for control of the assistive system. These findings indicate a substantial decoupling from operator-specific physical limitations, thereby enhancing accessibility for individuals with a broader range of motor impairments. Furthermore, the system’s independent, on-demand nature facilitates seamless integration into the operator’s daily activities, both within and beyond the home environment.

\subsection{At Home}

Overall, the successful application of our system in a real-world environment shows the utility and generalizability of the proposed system. Although we demonstrate performance only on a limited number of tasks, no fine-tuning was performed on them. From this, it stands to reason that the system has the potential to tackle a much wider variety of household tasks than presented. Additionally, due to the nature of the legged base, the system could also accompany the operator outside the household environment, for example, to visit a grocery store. The improvement in \emph{Mental Demand} and \emph{Effort} scores when introducing more autonomy into the system, in this case via the autonomous pre-positioning unit, confirms the intuition that improved machine intelligence reduces the mental load on the operator. We acknowledge that the ability to draw definitive conclusions from a single-participant user study is inherently limited. However, due to the substantial scaling challenges associated with operator training, extending the study to additional participants is non-trivial.

\subsection{Future Work}
In future work, we plan to expand the scope of this research to include additional participants. This will allow us to evaluate how well the system generalizes across a broader range of operators and disabilities, and to assess whether a control scheme developed in close collaboration with a single expert operator remains effective for new users.

Furthermore, a current limitation of the system is that its autonomy does not yet extend to generalized physical interaction with the environment. While the robot can navigate autonomously, it does not independently manipulate objects or perform physical tasks. The usefulness of an assistive robot increases with the number of tasks it can complete autonomously, i.e., without continuous operator oversight. Future developments will therefore focus on extending autonomy to interaction tasks, including more advanced pick-and-place behaviors, door opening, and assistive manipulation such as feeding. Additionally, related shared-autonomy approaches have proposed supported teleoperation methods that reduce the cognitive load associated with manual object manipulation~\cite{padmanabha2024independence, doi:10.1126/scirobotics.aeb6725}. Investigating the integration of such techniques into our system is a promising direction for future work.

\section{Conclusions}\label{sec:conclusion}


We presented the first design of an on-demand quadrupedal assistance robot intended for people with severe motor impairments. The proposed platform offers high versatility, reaching areas inaccessible to wheelchairs while freeing the operator from the need to permanently carry a robotic arm attachment. We demonstrated the system at the 2024 \emph{Cybathlon Assistance Robot Race}, where it achieved performance and time-efficiency comparable to wheelchair-mounted systems while offering greater operator autonomy. In a follow-up study, we further validated the system’s generalizability in a real-world home environment, where it successfully assisted the operator in completing a range of everyday tasks. The results highlight strong performance across key subjective metrics, including efficiency, error rate, and mental workload. Moreover, the rapid improvement and increasing availability of quadrupedal platforms, as well as emerging humanoid systems, underscore substantial opportunities for advancing assistive robotic technologies and expanding their applicability in the daily lives of people with limited mobility.

\backmatter

\section{Supplementary}\label{sec:sup}

\section*{Declarations}

\bmhead{Ethics approval and consent to participate}
Under guidance from the ETH Ethics Commission, the user study was framed as a contract of collaboration between the researchers and the participant. This was done to reflect both our collaboration in the development of the system for the competition itself and their role as an expert operator of the proposed system. The proposal detailing the user study was reviewed by the ETH Zurich Ethics Commission. Based on this review, the Vice President for Research of ETH Zurich, Prof. Annette Oxenius, approved it without reservations. The official proposal title and reference number were as follows: \emph{Project 25 ETHICS-236: Cybathlon-Inspired Mobile Robotic Assistance at Home}.

\bmhead{Consent for publication}
Consent was collected from the participant of the user study to gather data and video material containing them and their home. We explicitly reaffirmed their consent to have this data published in a non-anonymized form in the context of this publication. 

\bmhead{Availability of data and materials}
The robot data generated and analyzed during the at-home study will not be made publicly available due to privacy concerns of the operator. 
All other data can be provided upon reasonable written request to the authors.

\bmhead{Competing interests}
The authors declare that they have no competing interests.

\bmhead{Funding}
This project received funding from the internal ETH Zurich university budget. In addition, external funding for salaries happened through the European Union’s Horizon Europe Framework Program under grant agreement No 101121321, and from Huawei Tech R\&D (UK) through a research funding agreement. The external funding entities had no direct role in the project itself.

\bmhead{Authors' contributions}
CS and AC co-lead the RSL Cybathlon 2024 team. CS developed the \emph{Quadstick Interface}, as well as the \emph{Two-Robot Solution}, the \emph{Keypoint Map}, the \emph{Behavior Engine}, and the scaffolding of the at-home automation stack. She designed and led the at-home experiments, completed the ethics approval process, and adapted the autonomous navigation software of \emph{Waverider} for the at-home experiments. AC wrote the software for the localization of the robots on the Cybathlon racetrack. CC adapted the \emph{TagMap} infrastructure for the at-home experiments and was crucial in their execution. AC and JRC assisted with various hardware integration and inter-robot networking for Cybathlon 2024. MH provided the infrastructure and resources necessary for all the work presented and offered high-level guidance. CS composed the initial draft of the manuscript. All authors read, revised, and approved the final manuscript.

\bmhead{Acknowledgments}
First and foremost, we would like to express our deep gratitude to our expert operator, Samuel Kunz, without whom none of this work could have been possible. In the user study and especially during the long months of competition preparation, we could not have asked for a collaborator more eager to celebrate successes and more patient in dealing with system failures.
Furthermore, we would like to thank Yuni Fuchioka for his help at Cybathlon 2024, as well as Eris Sako and Fabian Tischhauser for assisting with hardware issues.
Finally, we thank Olga Vysotska for providing feedback on the early draft of the manuscript.





\newacronym{dof}{DoF}{Degrees of Freedom}
\newacronym{mpc}{MPC}{Model Predictive Control}
\newacronym{ee}{EE}{End-Effector}
\newacronym{gui}{GUI}{Graphical User Interface}
\newacronym{llm}{LLM}{Large Language Model}
\newacronym{slam}{SLAM}{Simultaneous Localization and Mapping}
\newacronym{cnn}{CNN}{Convolutional Neural Network}
\newacronym{dnf}{DNF}{Did Not Finish}

\printglossary[type=\acronymtype, title=List of Abbreviations, toctitle=List of Abbreviations]

\bibliographystyle{plain}
\bibliography{sn-bibliography}

\end{document}